\pdfoutput=1
\documentclass{article}

\usepackage{microtype}
\usepackage{graphicx}
\usepackage{subfigure}
\usepackage{booktabs} 

\usepackage{hyperref}

\usepackage[accepted]{icml2023}

\usepackage{amsmath}
\usepackage{amssymb}
\usepackage{mathtools}
\usepackage{amsthm}

\usepackage[capitalize,noabbrev]{cleveref}

\theoremstyle{plain}
\newtheorem{theorem}{Theorem}[section]

\newtheorem{lemma}[theorem]{Lemma}
\newtheorem{corollary}[theorem]{Corollary}
\theoremstyle{definition}
\newtheorem{definition}[theorem]{Definition}
\newtheorem{assumption}[theorem]{Assumption}
\theoremstyle{remark}
\newtheorem{remark}[theorem]{Remark}

\usepackage[textsize=tiny]{todonotes}
\newcommand{\yifangb}[1]{\todo[inline,color=blue!10]{\textbf{YF: }#1}}
\newcommand{\yipingb}[1]{\todo[inline,color=orange!10]{\textbf{YP: }#1}}

\newcommand{\tnu}{\widetilde{\nu}}

\newcommand{\usigma}{\underline{\sigma}}

\newcommand{\hB}{\hat{B}}
\newcommand{\hW}{\hat{W}}

\newcommand{\Bs}{B^*}
\newcommand{\Ws}{W^*}

\newcommand{\argmin}[1]{\mathop{\arg\min}\limits_{#1}}
\newcommand{\E}{\mathbb{E}}
\newcommand{\R}{\mathbb{R}}
\newcommand{\gradient}{\nabla}
\newcommand{\yiping}[1]{\textcolor{green}{[YP:~#1]}}

\newcommand{\yifang}[1]{\textcolor{blue}{[YF:~#1]}}
\newcommand{\kevin}[1]{\textcolor{magenta}{[KJ:~#1]}}

\icmltitlerunning{Improved Active Multi-Task Representation Learning via Lasso}

\begin{document}

\twocolumn[
\icmltitle{Improved Active Multi-Task Representation Learning via Lasso}



\begin{icmlauthorlist}
\icmlauthor{Yiping Wang}{zju}
\icmlauthor{Yifang Chen}{uw}
\icmlauthor{Kevin Jamieson}{uw}
\icmlauthor{Simon S. Du}{uw}
\end{icmlauthorlist}

\icmlaffiliation{zju}{College of Computer Science and Technology, Zhejiang University}
\icmlaffiliation{uw}{Paul G. Allen School of Computer Science \& Engineering, University of Washington}

\icmlcorrespondingauthor{Yiping Wang}{yipingw@zju.edu.cn}



\vskip 0.3in
]



\printAffiliationsAndNotice{} 


\begin{abstract}
To leverage the copious amount of data from source tasks and overcome the scarcity of the target task
samples, representation learning based on multi-task pretraining has become a standard approach
in many applications. However, up until now, most existing works design a source task selection strategy from a purely empirical perspective. Recently, \citet{chen2022active} gave the first active multi-task representation learning (A-MTRL) algorithm which adaptively samples from source tasks and can provably reduce the total sample complexity using the L2-regularized-target-source-relevance parameter $\nu^2$. But their work is theoretically suboptimal in terms of total source sample complexity and is less practical in some real-world scenarios where sparse training source task selection is desired. In this paper, we address both issues. Specifically, we show the strict dominance of the L1-regularized-relevance-based ($\nu^1$-based) strategy by giving a lower bound for the $\nu^2$-based strategy. 
When $\nu^1$ is unknown, we propose a practical algorithm that uses the LASSO program to estimate $\nu^1$.
Our algorithm successfully recovers the optimal result in the known case. In addition to our sample complexity results, we also characterize the potential of our $\nu^1$-based strategy in sample-cost-sensitive settings. Finally, we provide experiments on real-world computer vision datasets to illustrate the effectiveness of our proposed method.
\end{abstract}
\section{Introduction}
\label{introduction}

Deep learning has been successful because it can effectively learn
a proper feature extractor that can map high-dimensional, highly structured inputs like natural images and natural language into a relatively low-dimensional representation.   
Recently, a big focus in deep learning has been on few-shot learning, where there is not enough data to learn a good representation and a prediction function from scratch. One solution is using multi-task learning, which uses data from other sources to help the few-shot target. This approach is based on the idea that different tasks can share a common representation. The process starts by training on a lot of source tasks to learn a simpler representation and then uses that pre-trained representation to train on a limited amount of target data.
Accessing a large amount of source data for multi-task representation learning (MTRL) may be easy, but processing and training on all that data can be costly. Therefore, it is important to find ways to minimize the number of samples, and perhaps the number of sources, needed from source tasks while still achieving the desired performance on the target task. 
Naturally, not all source tasks are equally important for learning the representation and maximizing performance on the target task. 
But to the best of our knowledge, most research in this area chooses which tasks to include in the training of the multi-task representation in an ad hoc way~\citep{asai2022attentional,fifty2021efficiently,yao2022nlp,zaiem2021pretext,zamir2018taskonomy,zhang2022task}.
Notable exceptions include \cite{chensx2021weighted,chen2022active} that study ways to improve training efficiency and reduce the cost of processing source data by prioritizing certain tasks during training with theoretical guarantees.

On the other hand, the significant empirical success of MTRL has motivated a number of theoretical studies \cite{Du2021FewShotLV,chen2022active, Tripuraneni2021ProvableMO}. In particular, \cite{Du2021FewShotLV} and \cite{Tripuraneni2021ProvableMO} provide generalization (excess risk) upper bounds  on the estimation error of the target task for passive multi-task representation learning (P-MTRL). Here, \emph{passive} means that samples are drawn from tasks according to some non-adaptive sampling strategy fixed before data is observed (e.g., an equal number of samples from each task).
%
\citet{Tripuraneni2021ProvableMO} also proves a lower bound related to the quality of whole feature representations in P-MTRL.
 

In this paper, our main focus is to guarantee a specific level of accuracy on a target task while provably using the least amount of data from other related tasks. This is achieved through task-level active learning. 
\citet{chen2022active} is the first work to propose an active multi-task representation learning (A-MTRL) algorithm that can provably reduce the total number of samples from all the tasks compared to the passive learning version (P-MTRL) by estimating the relevance of each source task to the target task and sampling accordingly.
However, this previous work has several limitations and leaves some questions open, in both theory and practical application. For example, they did not study the lower bounds of the excess risk on the target task for Multi-Task Transfer Learning. 
Furthermore, \citet{chen2022active} proposed an $L_2$ regularized source-to-target-task relevance quantity $\nu^2$, but it is unclear whether this relevance score is the best criterion for the A-MTRL design compared to other possible relevance scores. As we will show later, their A-MTRL algorithm may not be optimal.
In our work, we build on \cite{chen2022active} by optimizing their upper bound of the excess risk and show that this yields an asymptotically optimal sampling strategy which corresponds to an $L_1$ regularized relevance quantity $\nu^1$ and samples from this distribution accordingly. 
Moreover, we provide the first sampling-algorithm-dependent 
minimax lower bound of excess risk on the target task for both the A-MTRL in \cite{chen2022active} and P-MTRL, which shows that our algorithm can strictly outperform these baselines even in the worst case.

In addition to the theoretical bounds, \citet{chen2022active} also has practical limitations. 
When there exist multiple sampling strategies that are seemingly equivalent under their framework, their algorithm tends to put a little weight on all tasks by nature of the $L_2$ regularized solution $\nu^2$. 
This is sometimes undesirable in practice as will 
 illustrate by two examples. First, setting up a sample-generating source can be more expensive than actually generating the samples. For example, in robotics, each source task can be considered as a specific real-world testing environment that can take weeks to set up, but then samples can be generated quickly and plentifully \citep{NEURIPS2021_52fc2aee}. Second, previous research assumes that the cost of samples is the same no matter how much data we need or for how long. However, in reality, subscribing to data from a single source for a long period of time can lead to a lower average label cost. Therefore, even with the same sample complexity from sources, choosing fewer source tasks can be more beneficial. We propose a general-purpose cost-sensitive A-MTRL strategy that addresses these scenarios and demonstrates the potential of our proposed $L_1$ regularized strategy in various
  cost-effective
 situations.

\subsection{Our Contributions}

We summarize our contributions as follows.
\begin{itemize}


\item We begin by proving that the sampling distribution over tasks using our proposed $L_1$ strategy defined in Def.~\ref{def_LpNq} minimizes the target excess risk upper bound of \cite{chen2022active}. 
We then consider a class of strategies Lp-A-MTRL (A-MTRL with $L_p$ strategy) and show that, when $T \gtrsim k^2$, for $N_{tot}$ number of total source samples, L1-A-MTRL is strictly dominant over this class by proving that its 
estimation error decreases at least as fast as $\mathcal{\widetilde{O}}(\frac{k}{\sigma_k^2 N_{tot}})$ while the error of the L2-A-MTRL/P-MTRL strategies suffers algorithm-dependent minimax lower bound of at least $\widetilde{\Omega}(\frac{T}{k\sigma_k^2N_{tot}})$. Here $T$ is the number of source tasks, $k$ is the dimension for the non-shared prediction function and $\sigma_k$ characterizes the diversity of source tasks which will be specified later.
These minimax lower bounds are novel to the MTRL literature. 
(Section~\ref{sec_knwon_nu}, \ref{sec: minimax bound}) 



\item While the L1-A-MTRL strategy provably has sample complexity benefits over other sampling strategies, it is not directly implementable in practice since it requires prior knowledge of $\nu^1$ (i.e., those bounds only demonstrate the performance of the sampling distribution, not how to find it). 
Consequently, inspired by \cite{chen2022active}, we design a practical strategy that utilizes the Lasso and a low order number of samples to estimate the relevance vector $\nu^1$, and then apply the $L_1$ strategy to sample source data using the estimated  $\nu^1$.
We show that this practical algorithm achieves a sample complexity nearly as good as when $\nu^1$ is known. The key technical innovation here is that when the regularization parameter is lower bounded, the Lasso solution can be close to the ground truth value. (Section~\ref{sec_P_L1_A_MTRL})

\item 
Going beyond these main results, we demonstrate that our L1-A-MTRL strategy can be extended to support many sample-cost-sensitive scenarios by levering its sparse source task selection properties. 
We formulate this setting as an optimization problem and formally characterize the benign cost function under which our L1-A-MTRL strategy is beneficial (Section~\ref{sec_task_selection})
\item 
Finally, we empirically show the effectiveness of our algorithms. 
If we denote the practical algorithm of \cite{chen2022active} by L2-A-MTRL, we show that our proposed 
L1-A-MTRL algorithm  achieves 0.54\% higher average accuracy on MNIST-C 
relative to L2-A-MTRL ($92.6\%$), which confirms our theoretical results.
We then restrict most of the data to be sampled from no more than 10 tasks, in order to mimic the sample-cost-sensitive setting with decreasing per-sample cost. Here we 
find L1-A-MTRL achieves 2.2\% higher average accuracy relative to the uniform sampling ($94.3\%$).  (Section~\ref{sec_experiment}).
\end{itemize}


\section{Preliminaries}
In this section, We describe the relevant notations and the problem setup for further theoretical analysis.

\subsection{Notation}\label{sec_notation}
\textbf{Miscellaneous.} Let $[T]:= \{1 ,2,...,T\}$ denotes the set of source tasks and  
$n_{[T]}:= \{n_1, n_2, ..., n_T\}$ denotes the number of samples dedicated to each task. 
Likewise, $n_{[T],i}:= \{n_{1,i},...,n_{T,i}\}$ represents the data from each task at the $i$-stage for multi-stage learning procedure.
If $S$ is an index set, $|S|$ denotes the number of elements in $S$. 
We use $\Vert \cdot\Vert_p$ to denote the $l_p$ norm of vectors and use $|\cdot | $ or $ \|\cdot\|$ to denote the $l_2$ norm for convenience. 
Let subG$_{d}(\rho^2)$ be the $d$-dimensional sub-gaussian variables with variance $\rho$. 

\noindent
\textbf{Singular Values.} 
For $A \in \R^{m\times n}$, we denote by $\sigma_i(A)$ the $i$-th singular value of $A$, which satisfy $\sigma_1(A)\geq ... \geq \sigma_r(A) > 0$ with $r = rank(A)$. And we specify $\kappa(A)$ as the condition number of $A$, i.e., $\kappa(A) = \frac{\sigma_1(A)}{\sigma_r(A)}$ if $\sigma_r(A) > 0$. 

\noindent
\textbf{Asymptotic comparison.} We use the standard $O, \Omega, \Theta$ notations to hide the universal constants, and further use $\widetilde{O}, \widetilde{\Omega}, \widetilde{\Theta}$ to hide logarithmic factors. We use $a \lesssim b$ or $b \gtrsim a$ to denote $a = O(b)$ and use $a \asymp b$ to denote $a = \Theta(b)$.



\subsection{Problem Setup}\label{sec_setup}
\textbf{Multi-Task.} Let $t \in [T]$ be the index of the $T$ source tasks and index $T+1$ denotes the target task. Each task $t \in [T+1]$ is associated with a joint distribution $\mu_t$ over $\mathcal{X}\times \mathcal{Y}$, where $\mathcal{X}$ is the input space and $\mathcal{Y}$ is the output space. In this paper we assume $\mathcal{X} \subseteq \R^{d}$ and $\mathcal{Y} \subseteq \R$. 

\noindent
\textbf{Data Generation.} Like in \cite{chen2022active}, we assume there exists an underlying representation function $\phi^*: \mathcal{X} \rightarrow \mathcal{R}$ which maps the input space $\mathcal{X}$ to a feature space $\mathcal{R} \in \R^k$ where $k \ll d$. And the representation functions are restricted to be in some function classes $\Phi$, e.g., linear functions, convolutional networks, etc. We further assume that each $t$-th task for $t \in [T+1]$ follows a ground truth linear head $w_t^*$ that maps the particular feature to the corresponding label. To be more specific, we assume the \textit{i.i.d} sample $(x_t,y_t) \sim \mu_t$ satisfies
\begin{equation}\label{equ_data_gen}
    y_t = \phi^*(x_t)^\top w_t^* + z_t, \quad z_t \sim \mathcal{N}(0, \sigma_z^2)
\end{equation}
where $x_t \sim p_t$ and $x_t$ is independent to $z_t$. 
For convenience, we denote $X_t = [x_{t,1}, .., x_{t,n_t}]^\top \in \R^{n_t \times d}$ to be the input data matrix which contained $n_t$ \textit{i.i.d.} sampled data $(x_{t,1}, y_{t,1}),...,(x_{t,n_t}, y_{t,n_t}) \sim \mu_t$ from the $t$-th task, and $Y_t = [y_{t,1}, ..., y_{t,n_t}]^\top \in \R^{n_t}$, $Z_t = [z_{t,1}, ..., z_{t,n_t}] \in \R^{n_t}$ to be the labels and noise terms aligned to the inputs. For convenience, we define $N_{tot} := \sum_{t=1}^Tn_t$ to be the total sampling number from all the source tasks.

\noindent
\textbf{Transfer Learning Process.}
As in \cite{Du2021FewShotLV}, firstly we learn the representation map on the source tasks by solving the following optimization problem
\begin{equation}\label{equ_train_on_source}
    \hat\phi, \hat{w}_1, ..., \hat{w}_T = \argmin{\phi\in\Phi, w_1,...,w_T \in \R^k} \frac{1}{T}\sum_{t=1}^T \frac{1}{n_t}\Vert Y_t - \phi(X_t) w_t\Vert_2^2
\end{equation}
Here we allow $n_t$ to vary from different tasks rather than requiring uniform sampling and $\phi(X_t):= [\phi(x_{t,1}), ..., \phi(x_{t,n_t})]^\top \in \R^{n_t\times k}$. Then we retain the learned representation and apply it to the target task while training a specific linear head for this task:
\begin{equation}\label{equ_train_on_target}
    \hat{w}_{T+1} = \argmin{w_{T+1}} \frac{1}{n_{T+1}}\Vert Y_{T+1} - \hat{\phi}(X_{T+1})w_{T+1}\Vert_2^2
\end{equation}
\noindent
\textbf{Goal.} Our main goal is to bound the \textit{excess risk} (ER) of our model on the target task with parameters $(\hat{\phi}(x), \hat{w}_{T+1})$ while minimizing the total cost of sampling data from the source tasks.
Here, like in \cite{Du2021FewShotLV, chen2022active}, we define the population loss as $L_{T+1}(\hat{\phi}, \hat{w}_{T+1}) = \mathbb{E}_{(x,y)\sim \mu_{T+1}}[(y_{T+1}-\hat{\phi} (x_{T+1})^\top \hat{w}_{T+1})^2]$.
Then from (\ref{equ_data_gen}) we can define the excess risk:
\begin{equation}\label{equ_excess_risk}
\begin{split}
    &ER(\hat{\phi}, \hat{w}_{T+1}, \phi^*, w_{T+1}^*)\\ &= L_{T+1}(\hat{\phi}, \hat{w}_{T+1}) - L_{T+1}(\phi^*, w^*_{T+1})\\
    &= \mathbb{E}_{x\sim p_{T+1}}[(\hat{\phi}(x)^\top \hat{w_t} - \phi^*(x)^\top w_t^* )^2]
\end{split}
\end{equation}
It should be mentioned that in this paper, we consider the model performance under the worst circumstance, therefore we treat the ground truth parameters $\phi^*, w_{T+1}^*$ as the arguments of excess risk, which is different from that in the previous works~\cite{Du2021FewShotLV, chen2022active}. 

\textbf{Linear Representation.} Our theoretical study concentrates on the linear representation function class, which is widely used in many previous works \cite{Du2021FewShotLV, tripuraneni2020theory, Tripuraneni2021ProvableMO, thekumparampil2021sample,chen2022active}. Namely, we let $\Phi = \{x\mapsto B^\top x \ | \ B \in \R^{d\times k}\}$ and thus $\phi(X_t) = X_tB \in \R^{n_t\times k}$. Without loss of generality, we assume the ground truth representation map $B^*$ is an orthonormal matrix, i.e., $B^* \in O_{d,k}$, which is also commonly used in the related works \cite{chen2022active, Tripuraneni2021ProvableMO, kumar2022fine}. 

\vspace{3pt}
\textbf{Other assumptions.} 
Assume $\E_{x_t\sim p_t}[x_t] = 0$, $\Sigma_t^* = \E_{x_t\sim p_t}[x_t x_t^\top]$ and $\hat{\Sigma}_{t} := \frac{1}{n_t}(X_t)^\top X_t$ for any $t \in [T+1]$. We have the following assumptions for the data distribution:
\begin{assumption}\label{assump_subG} (sub-gaussian input).
    There exists $\rho \geq 1$ such that $x_t \sim p_t$ is subG$_{d}(\rho^2)$ for all $t \in [T+1]$. 
\end{assumption}
\begin{assumption}\label{assump_theta1_var}
    (proper variance) 
    For all $t \in [T+1]$, we have $\sigma_{\max}(\Sigma_t^*) = \Theta(1)$ and $\sigma_{\min}(\Sigma_t^*) = \Theta(1)$.
\end{assumption}
Variance conditions are common in the related works \cite{Tripuraneni2021ProvableMO, Du2021FewShotLV, chen2022active}
and Assumption~\ref{assump_theta1_var} is a generalization than identical variance assumption used in \cite{Tripuraneni2021ProvableMO, chen2022active} which requires $\Sigma_1 = ... = \Sigma_{T+1} = I_d$. Specially, we only use the identical variance assumption in Section~\ref{sec_P_L1_A_MTRL}.
\begin{assumption}\label{assump_dimension}
    (high dimension input and enough tasks) The parameters satisfy $d > T \geq k \geq 1$ and $d \gg k$.
\end{assumption}

Finally, we also need \textit{diverse task} assumption mentioned in \cite{Tripuraneni2021ProvableMO, Du2021FewShotLV, chen2022active}. Denote $W^*:= [w_1^*,...,w_T^*] \in \R^{k\times T}$, then we assume:

\begin{assumption}\label{assump_W_fullr}
    (diverse task) The matrix  $W^*$ satisfies $\sigma_{\min}(W^*) > 0$.
\end{assumption}

Assumption~\ref{assump_W_fullr} claims that $W^*$ has full row rank, so we can definitely find some $\nu \in \R^{T}$ such that $W^*\nu = w_{T+1}^*$, and thus (\ref{equ_Lp_min}) in Def.~\ref{def_LpNq} is well-defined. It's a necessary assumption for learning reasonable  features as proven by \cite{Tripuraneni2021ProvableMO}.

\subsection{Scope of A-MTRL algorithms in this paper}

Here we state the scope of the A-MTRL algorithm considered in this paper. 
Recall that in \cite{chen2022active}, the learner samples in proportional to $\frac{\hat{\nu}(t)^2}{\|\hat{\nu}\|_2^2}$ number of data from task $t$, where $\hat{\nu}$ is defined via the following solution:
\begin{equation}\label{equ_Lp_min}
    \arg \min_{\nu}\Vert \nu\Vert_2 \qquad \text{s.t.} \  W^*\nu = w^*_{T+1}
\end{equation}
Here instead of focusing on this $L_2$ regularization, we study the whole candidate set of source-target relevance terms and the corresponding sampling strategies. Formally, we generalize Definition 3.1 of \cite{chen2022active} to propose:
\begin{definition}\label{def_LpNq} ($L_pN_q$ sampling strategy)~ Let $\nu(t)$ be the $t$-th element of vector $\nu\in\R^T$ and $\underline{N}$ be the minimum number of sampling data from every source task. The $L_pN_q$ strategy is defined as taking $n_t = \max\{c^\prime |\nu^p(t)|^q, \underline{N}\}$ for some constant $c^\prime>0$, where $n_t$ is the number of samples drawn from from the $t$-th task, and  
\begin{equation}\label{equ_Lp_min}
    \nu^p = \arg \min_{\nu}\Vert \nu\Vert_p \qquad \text{s.t.} \  W^*\nu = w^*_{T+1}.
\end{equation}
If $p=q$, we denote $L_p$ as the abbreviation of $L_pN_q$. For example, if $\underline{N} = 0$, then the $L_1$ strategy corresponds to $n_t = \frac{N_{tot}}{\Vert\nu^1\Vert_1}\cdot |\nu^1(t)|$ and the $L_2$ strategy corresponds to $n_t = \frac{N_{tot}}{\Vert\nu^2\Vert^2_2} |\nu^2(t)|^2$, where $N_{tot}$ is the total source sampling budget.
%
\end{definition}

In the rest of the paper, we will focus on this $L_pN_q$ sampling strategy set.
\section{Main Results}
\label{sec_main_results}


\subsection{Optimal Strategy L1-A-MTRL with Known $\nu$}
\label{sec_knwon_nu}
In this section, we aim to obtain the optimal sampling strategy that can achieve the required performance on the target task with the smallest possible number of samples from source tasks.
Firstly, with linear representation assumption, we rewrite $ER(\hat{B},\hat{w}_{T+1},B^*,w_{T+1}^*)$ in (\ref{equ_excess_risk})  as follows:
\begin{equation}\label{equ_def_ori_ER}
\E_{x \sim p_{T+1}} \Vert x^\top(\hat{B}\hat{w}_{T+1} - B^*w_{T+1}^*) \Vert_2^2.
\end{equation}
Then from the intermediate  result of Theorem 3.2 in \cite{chen2022active}, we get the upper bound of excess risk for all A-MTRL methods:
\begin{theorem} \label{thm_chen_UB}\cite{chen2022active}
Fix a failure probability $\delta \in (0, 1)$. If Assumption~\ref{assump_subG}, \ref{assump_theta1_var}, \ref{assump_dimension}, \ref{assump_W_fullr} hold,  
and the sample size in source and target tasks satisfy $n_t \gg\rho^4(d + \ln(\frac{T}{\delta}))$ for all $t\in[T]$ and $n_{T+1}\gg\rho^4(k + \ln(\frac{1}{\delta}))$, 
then with probability at least $1-\delta$ we have:
\begin{equation} 
\label{inequ_ER_upper_wildenu}
\begin{aligned}
    &ER(\hat{B}, \hat{w}_{T+1}, B^*, w_{T+1}^*) 
    \\&\lesssim \sigma^2(kd\ln(\frac{N_{tot}}{T}) + kT +\ln(\frac{1}{\delta}))\Vert\widetilde{\nu}\Vert_2^2+ \sigma^2\frac{(k +  \ln(\frac{1}{\delta}))}{n_{T+1}}
\end{aligned}
\end{equation}
where $\nu \in \{\nu'\in\R^T|W^*\nu' = w_{T+1}^*\}$
and $\tnu(t) = \frac{\nu(t)}{\sqrt{n_t}}$.
\end{theorem}

The key idea behind Theorem~\ref{thm_chen_UB} is as follows.
\cite{Du2021FewShotLV} provides the first upper bound for the MTRL problem. They consider sampling data evenly from each source task and demonstrated that following the transfer learning process (Eqn.~\ref{equ_train_on_source},~\ref{equ_train_on_target}), the  target task error can be controlled by the source-task training error $\mathcal{O}(\sigma^2(kd+kT)/{N_{tot}})$ and the target-task fine-tuning error $\mathcal{O}( \sigma^2 k/n_{T+1})$. However, in their proof, the ground truth linear head $w_{T+1}^*$ is required to satisfy a distribution $Q$ such that $\Vert E_{w\sim Q}[ww^\top]\Vert \leq O(\frac{1}{k})$.
\cite{chen2022active} go beyond this limitation by leveraging the equation $W^*\nu^*=\widetilde{W}^*\widetilde{\nu}^* = w^*_{T+1}$, where $\widetilde{w}^*_t=w^*_t \sqrt{n_t}$ and $\widetilde{\nu}^*(t) = \frac{\nu^*(t)}{\sqrt{n_t}}$. This idea introduces the source-target relevance vector $\nu^*\in R^T$ into the bound and results in Eqn.~\ref{inequ_ER_upper_wildenu}.

Inspired by Theorem~\ref{thm_chen_UB}, in order to minimize the excess risk bound  
with a fixed sampling quota $N_{tot}$, we need to find the optimal sampling strategy $n_{[T]}=\{n_1, ..., n_{T}\}$ by solving the following optimization problem:
\begin{equation}
\begin{aligned}\label{opt_min_nu_general}
    &\min_{\nu, n_{[T]}} \|\widetilde{\nu}\|_2^2 = \sum_{t=1}^T\frac{(\nu(t))^2}{n_t}\\
    \text{s.t. }& 
     W^* \nu = w_{T+1}^* \\
     &\sum_{t=1}^T n_t = N_\text{tot}\\
     &n_t \geq \underline{N},
     \quad \forall t \in [T]
\end{aligned}
\end{equation}
Here $\underline{N} (\gg \rho^4(d+\ln(\frac{T}{\delta})))$ is the minimum sampling number for every source task as in Theorem~\ref{thm_chen_UB}. In this section, we will transform (\ref{opt_min_nu_general}) into a bi-level optimization problem and obtain the asymptotic optimal solutions of (\ref{opt_min_nu_general}).

\subsubsection{Optimal Strategy for Any Fixed $\nu$}
We first consider a fixed $\nu$ in (\ref{opt_min_nu_general}) and find the optimal sampling strategy accordingly, and we get:
\begin{lemma}\label{lemma_nu}
For any fixed $\nu$ such that $W^*\nu = w_{T+1}^*$, the optimal $n^*_{[T]}$ for minimizing $\Vert\tnu\Vert_2^2$
satisfies
$n^*_{t} = \max\{c^\prime |\nu(t)|, \underline{N}\}$ for every $t \in [T]$, where $c^\prime > 0$ is some constant such that $\sum_{t=1}^Tn^*_t = N_{tot}$.
\end{lemma}
This lemma indicates an optimal sampling strategy under some fixed,  arbitrary $\nu \in \{ \nu' | W^* \nu' = w_{T+1}^*\}$.
We can then apply Lemma~\ref{lemma_nu} to the previous bound (\ref{inequ_ER_upper_wildenu}) and deduce the theoretical optimal bound on the sample complexity of the source tasks for any suitable $\nu$. Here, for simplicity, we skip the trivial case where the model achieves sufficiently high accuracy with uniformly allocated sampling data $\underline{N}$ by requiring $\varepsilon^2 \ll \min(1, \sigma^2(kd+kT)\frac{\Vert \nu\Vert_1^2}{T\underline{N}})$. This condition guarantees that $N_{tot}\gg T\underline{N}$, and we get:

\begin{corollary}
    \label{corollary_sample_nu_known}
    Assume Assumption~\ref{assump_subG}, \ref{assump_theta1_var}, \ref{assump_dimension}, \ref{assump_W_fullr} hold and $\nu$ is fixed. Then the optimal sampling strategy $n_{[T]}$ satisfies $n_{t} = \max\{c^{\prime} |\nu(t)|, \underline{N}\}, ~\forall t \in [T]$, and with probability at least $1-\delta$, the optimal A-MTRL algorithm satisfies $ER \leq \varepsilon^2$ with $\varepsilon^2 \ll \min(1, \sigma^2(kd+kT)\frac{\Vert \nu\Vert_1^2}{T\underline{N}})$ whenever the total sampling budget from all source tasks $N_{tot}$ is at least
    \begin{equation}
        \widetilde{\mathcal{O}}(\sigma^2(kd+kT)\Vert \nu\Vert_1^2\varepsilon^{-2})
    \end{equation}
    and the number of target samples is at least $~\widetilde{\mathcal{O}}(\sigma^2k\varepsilon^{-2})$. 
\end{corollary}

\textbf{Discussion.}
To show the optimality of our bound, we compare this with the result in \cite{chen2022active}. Their known $\nu^2$ (denoted as $\nu^*$ in their original paper) is equivalent to 
$$
    \arg\min_{\nu} \|\nu\|_2 \quad\text{ s.t.} ~W^*\nu = w_{T+1}^*.
$$
Under the same setting but using this $\nu^2$,
with probability at least $1-\delta$ , A-MTRL algorithm with sampling strategy $n_{[T]}$ such that $n_{t} = \max\{c^{\prime\prime} (\nu(t))^2, \underline{N}\}, ~\forall t \in [T]$ satisfies $ER \leq \varepsilon^2$  with $\varepsilon \ll 1$ whenever $N_{tot}$ is at least 
    \begin{equation}
        \widetilde{\mathcal{O}}(\sigma^2(kd+kT)s^*\Vert \nu^2\Vert_2^2\varepsilon^{-2})
    \end{equation}
    and the required number of target samples is also $~\widetilde{\mathcal{O}}(\sigma^2k\varepsilon^{-2})$. Here $s^* = \min_{\gamma \in [0, 1]}(1-\gamma)\Vert\nu^2 \Vert_{0, \gamma}+\gamma T$ and $\Vert\nu^2 \Vert_{0, \gamma}:= |\{t: |\nu^2(t)| > \sqrt{\gamma \Vert \nu^2\Vert_2^2 N_{tot}^{-1}}\}|$. 
    From Lemma~\ref{lemma_nu} we know our strategy is better than the previous under given arbitrary $\nu$ setting, so we have $\Vert\nu\Vert_1 \lesssim \sqrt{s^*}\Vert\nu\Vert_2 \leq \sqrt{T}\Vert\nu\Vert_2, \forall \nu \in \{\nu'|W^*\nu'=w_{T+1}^*\}$. In particular, we show the gap between $\Vert\nu\Vert_1$ and $ \sqrt{s^*}\Vert\nu\Vert_2$ can be very large under some special cases as follows.
    

\noindent{\rm\textbf{Example: Almost Sparse $\nu$.}}
Assume $T \gg 1$, $N_{tot} \gg \underline{N}T \geq T$, then we consider an extreme case where
\begin{equation}
    \nu(t) = \left\{
    \begin{array}{cc}
    \sqrt{1-\frac{1}{T-1}}&, t = 1 \\
    \frac{1}{T-1}&, t = 2, ..., T \\
    \end{array}
    \right.
\end{equation}
Then $\nu$ is approximately 1-sparse since $\frac{1}{T-1}\ll 1$, and we have $\Vert \nu \Vert_1 = \sqrt{1-\frac{1}{T-1}} + 1 < 2$, $\Vert \nu \Vert_2 = 1$. Let $\gamma_0 := \frac{N_{tot}}{(T-1)^2}$, it's easy to prove $s^* \geq \min\{\gamma_0, 1\}\times T \gg 1$. This result in $\sqrt{s^*}\Vert \nu\Vert_2 \gg \Vert \nu\Vert_1$ and A-MTRL in \cite{chen2022active} requires a sample complexity that is $\min\{\frac{N_{tot}}{2(T-1)}, \frac{T}{2}\}$ times larger than our optimal sampling strategy.

\subsubsection{Optimal $\nu$ in Candidate Set}\label{sec_L1_worst_target_task}
Secondly, suppose we are able to access the whole set $\{\nu^\prime | W^*\nu^\prime = w_{T+1}^*\}$, now we aim to find the optimal $\nu$ from the candidate set for sampling. Once we find such a $\nu^*$, we can utilize rules in Lemma~\ref{lemma_nu} to obtain $n_{[T]}^*$ and apply all the results above. Here we focus on the case in \cite{chen2022active} where ER bound $\varepsilon^2\rightarrow 0$ and $N_{tot}\rightarrow +\infty$ and we deduce 
that $L_1$-minimization solution is the best choice.

\begin{theorem}\label{thm_optimal_sample_general}
    Let $(\nu^1, n^1_{[T]})$ denotes the sampling parameters of $L_1$ strategy defined in Def.~\ref{def_LpNq}, i.e.,
    \begin{equation}
    \begin{aligned}
        \nu^1 &= \arg \min_{\nu}\Vert \nu\Vert_1 \qquad \text{s.t.} \  W^*\nu = w^*_{T+1}\\
        n^1_{t} &= \max\{c^\prime |\nu^1(t)|, \underline{N}\}, \qquad \forall t \in [T]
    \end{aligned}
    \end{equation}
    Let $(\nu^*, n^*_{[T]})$ denote the optimal solution of (\ref{opt_min_nu_general}). Then as $N_{tot}\rightarrow +\infty$ we have $\nu^1 \rightarrow \nu^*$, $n^1_{[T]} \rightarrow n^*_{[T]}$. 
\end{theorem}


Theorem~\ref{thm_optimal_sample_general} shows that the $L_1$ strategy can correspond to the asymptotic optimal solution of (\ref{opt_min_nu_general}).
As a reference, Alg. 1 in \cite{chen2022active} is equivalent to $L_2$ strategy, and we call these classes of methods \textbf{Lp-A-MTRL} (A-MTRL with $L_p$ strategy) method with known $\nu^p$ for further discussion.

\subsection{How Good Is L1-A-MTRL with Known $\nu$? Comparison on the Worst Target Task}
\label{sec: minimax bound}

To show the effectiveness of the $L_1$ strategy with known $\nu^1$, 
we analyze the performance of MTRL algorithms on a worst-case target task $w_{T+1}^*$ that maximizes the excess risk.
Firstly, for better comparison, we define the sampling-algorithm-dependent minimax lower bound of excess risk. Let $\Gamma(\sigma_k)=\{W\in\R^{k\times T}| \sigma_{\min}(W) \geq \sigma_k\}$ 
for any $\sigma_k > 0$, then we define:
\begin{definition}\label{def_minimax_LB}(minimax ER lower bound) The mini-max lower bound of ER on the target task for Lp-A-MTRL method $\underline{ER}_{L_p}(\sigma_k)$ is defined as
\begin{equation}\label{equ_def_minimax_LB}
\begin{aligned}
    \inf_{(\hB, \hat{w}_{T+1})} \sup_{(\Bs, \Ws, w_{T+1}^*)} \E_{x \sim \mu_{T+1}} \Vert x^\top(\hat{B}\hat{w}_{T+1} - B^*w^*_{T+1}) \Vert_2^2\\
    =\inf_{(\hB, \hat{W})} \sup_{(\Bs, \Ws, \nu^p)} \E_{x \sim \mu_{T+1}} \Vert x^\top(\hat{B}\hat{W}\nu^p - B^*W^*\nu^p) \Vert_2^2
\end{aligned}
\end{equation}
where $W^*$ varies on $\Gamma(\sigma_k)$ such that Assumption~\ref{assump_W_fullr} holds and $\nu^p$ denotes the $L_p$-minimization solution of $W^*\nu = w_{T+1}^*$ like (\ref{equ_Lp_min}). Similar definitions hold for P-MTRL.
\end{definition}
\begin{remark}\label{remark_algo_minimax_LB}
    The left term of (\ref{equ_def_minimax_LB}) denotes the case that we consider the average error of the best prediction model $(\hat{B}, \hat{w}_{T+1})$ on any target task facing any possible ground truth parameters $(B^*, W^*, w_{T+1}^*)$. 
    The equality of (\ref{equ_def_minimax_LB}) holds because choosing $\hat{w}_{T+1}$ is equivalent to choosing any $\hat{W} \in \{W' | W'\nu^p = \hat{w}_{T+1}\}$, given the $(W^*,w_{T+1}^*)$-dependent $\nu^p$.
    Note that we consider the $L_p$ strategy as Def.~\ref{def_LpNq} which is determined by $(\nu^p, n_t)$, so once we choose some Lp-A-MTRL algorithm, (\ref{equ_def_minimax_LB}) just depends on  model parameters and $\sigma_k$.
\end{remark}

With this definition, we show that with known $\nu^p$, the ER on the worst target task for $L_1$-A-MTRL can reduce up to $\frac{T}{k}$ times of total sampling data from source tasks than that of $L_2$-A-MTRL\cite{chen2022active} and P-MTRL.

\begin{theorem}\label{thm_worst_target_ER}
Assume conditions in Theorem~\ref{thm_chen_UB} hold, $\|w_{T+1}^*\|=\Theta(1)$ and $\nu^1, \nu^2$ defined in Def.~\ref{def_LpNq} are known. Then for L1-A-MTRL, we claim $\nu^1$ is at most $k$-sparse, i.e., $\Vert\nu^1\Vert_0\leq k$. If $N_{tot}\gg T\underline{N}$ and $W^* \in \Gamma(\sigma_k)$, then with probability at least $1-\delta$, for ER defined in (\ref{equ_def_ori_ER}) we have
\footnote{
For the previous upper bound in Theorem~\ref{thm_chen_UB}, people estimate non-shared $w_{T+1}^*$ by linear-probing on the target task so (\ref{inequ_ER_upper_wildenu}) contains target-related error term. However, under the ''cheating'' case in Theorem~\ref{thm_worst_target_ER}, knowing $\nu^p$ means we already have such information as long as $n_t$ is large enough since $W^*\nu^p = w_{T+1}^*$. We want to emphasize that this known $\nu^p$ assumption is used for illustrating why $L_1$ strategy is better, but not for practical use.
}
:
\begin{equation*}
\begin{split}\label{equ_active_upper}
ER_{L_1} &\lesssim \sigma^2(kd\ln(\frac{N_{tot}}{T}) + kT+\ln(\frac{1}{\delta}) )\frac{k}{\sigma_k^2\cdot N_{tot}} 
    \end{split}
\end{equation*}
but for P-MTRL and $L_2$-A-MTRL, 
with probability at least $1-\delta$ we have 
:
\begin{equation*}
\begin{split}\label{equ_passive_lower}
    \underline{ER}_{L_2}(\sigma_k), \underline{ER}_{passive}(\sigma_k)
    \gtrsim  \sigma^2 \frac{dT}{\sigma_k^2\cdot N_{tot}} 
    \end{split}
\end{equation*}
So when $T \gtrsim k^2$, L1-A-MTRL outperforms L2-A-MTRL and P-MTRL for the worst target task.
\end{theorem}

\textbf{Discussion.} 
In essence, the sparsity of $\nu^p$ causes the difference in model performance on the worst-case target task. For the upper bound of L1-A-MTRL,
We show $\|\widetilde{\nu}^1\|_2^2 \lesssim k/(\sigma_k^2 \cdot N_{tot})$. And for the lower bound of L2-A-MTRL and P-MTRL, we utilize the fact that 
$\inf_{\hat{B},\hat{W}}\sup_{B^*,W^*}\|X_t(\hat{B}\hat{W} - B^*W^*)\|_2^2 \gtrsim \sigma^2kd$ (up to logarithmic factors) and the result that when the row of $\widetilde{W}^*$ is well aligned with $\widetilde\nu^2$, then $\|(\hat B - B^*)\widetilde W^* \widetilde \nu^2\| \gtrsim \|(\hat B - B^*)\widetilde W^*\|_F \|\widetilde \nu^2\|$, where $\nu^2$ can be chosen to satisfy $\|\widetilde \nu^2\| \gtrsim T/(k\cdot \sigma_k^2 \cdot N_{tot})$. 

\subsection{L1-A-MTRL Algorithm and Theory}\label{sec_P_L1_A_MTRL}

In the previous sections, we showed the advantage of A-MTRL with the $L_1$ sampling strategy when $\nu^1$ is given. However, in practice, $\nu^1$ is unknown and needs to be estimated from $W^*$ and $w_{T+1}^*$, which themselves need to be estimated with unknown representation $B^*$ at the same time. In this section, we design a practical L1-A-MTRL algorithm shown in \cref{alg:L1_A_MTRL} which  
estimates the model parameters $\hat{B},\hat{W}, \hat{w}_{T+1}$ and relevance vector $\hat{\nu}^1$. Here in our algorithm setting, 
we let 
\begin{equation}\label{equ_beta}
    \begin{aligned}
    \beta_1 &= 10^5  T k^{3}\cdot\frac{C_{W}^6}{\underline{\sigma}^{6}}(d+\ln(\frac{4T}{\delta}))\\
    \beta_2 &= k(d+T+\ln(\frac{1}{\delta}))\|\hat{\nu}^1\|_1^2 \varepsilon^{-2}+ \beta_1\\
    \end{aligned}
\end{equation}
where $C_W$ is defined in Assumption~\ref{assump_bounded_W}.
$\beta_1$ and $\beta_2$  characterize the sample complexity required to explore at the first and second stage, respectively, and they are determined by $T\beta$ and $N_{tot}$ defined in Theorem~\ref{thm_noise}.

We want to highlight that unlike the $L_2$-minimization approach of \cite{chen2022active}, our $L_1$-minimization solution does not have a closed form solution which creates more challenges in controlling the estimation error between $\hat{\nu}^1$ and $\nu^1$. To deal with this problem, we use the \textit{Lasso Program} 
~\cite{wainwright_2019,Tibshirani1996RegressionSA} to estimate $\hat{\nu}^1$:
\begin{equation}\label{equ_Lasso}
    \hat{\nu}^1 \in \arg \min_{\nu \in \R^T} \{\frac{1}{2}\Vert\hat{w}_{T+1} - \hat{W}\nu\Vert_2^2+\lambda_k \Vert \nu\Vert_1\}
\end{equation}
where the regularization parameter $\lambda_k$ is chosen by users. We prove that with proper $\lambda_k$, $\hat{\nu}^1$ will be sufficiently close to $\nu^1$ in $l_1$ norm when the following assumptions hold.

\begin{assumption}\label{assump_bounded_W}
(bounded norm) 
There exists $C_{W}, R > 0$ \text{s.t.} 
$\sigma_{\max}(W^*) \leq C_W$
and $\|w_{T+1}^*\|_2 =  \Theta (R)$.
\end{assumption}



\begin{assumption}(identical covariance)\label{assump_Sigma}
we have: $\Sigma_t = \Sigma^* = I_{d}$ for all $t \in [T+1]$.
\end{assumption}
Assumption~\ref{assump_bounded_W} implies $\forall t \in [T]$, $\Vert w_t^*\Vert_2 = \Vert W^*e_n\Vert_2 \leq C_W$, which is a very common condition in the previous work \cite{Du2021FewShotLV, Tripuraneni2021ProvableMO, chen2022active}.
Assumption~\ref{assump_Sigma} is a stronger variance condition than Assumption~\ref{assump_theta1_var}, but it's also used in \cite{Tripuraneni2021ProvableMO, chen2022active} and we only need it in this section.
With these assumptions we are prepared to state our theoretical guarantee for our practical L1-A-MTRL algorithm:
\begin{theorem}\label{thm_noise}
Let Assumption
~\ref{assump_subG},
\ref{assump_dimension}, \ref{assump_W_fullr},
\ref{assump_bounded_W},
\ref{assump_Sigma} 
hold. 
Let  $\gamma = \max\{2160k^{3/2}C_W^2/\usigma, \sqrt{2160k^{3/2}C_W^3/{\usigma}}\}$, where $\underline{\sigma} = \sigma_{\min}(W^*) >0$. For L1-A-MTRL method, we set the regularization parameter by:
\begin{equation}\label{equ_regular_init}
    \lambda_k = 45 \frac{\sqrt{k}RC_W\usigma}{\gamma}\max\{1,\frac{C_W}{\gamma}\}
\end{equation}
Then to let $ER_{L_1}\leq \varepsilon^2$ where $\varepsilon^2 \ll \min(1, \sigma^2(kd+kT)\frac{\Vert \nu\Vert_1^2}{T\underline{N}})$ with probability $1-\delta$, the number of source samples $N_{tot}$   is at most
\begin{equation}\label{equ_O_order}
\mathcal{\widetilde{O}}(\sigma^2(kd + kT )\Vert\nu^1\Vert_1^2\varepsilon^{-2} + T\beta)
\end{equation}
where $\beta = \max\{\gamma^2\frac{\sigma_z^2}{\underline{\sigma}^4},\gamma^2\frac{C_W^2}{\underline{\sigma}^4}\rho^4,\rho^4, \frac{\sigma_z^2}{\underline{\sigma}^2}\}\cdot(d + \ln(\frac{4T}{\delta}))$, and target task sample complexity $n_{T+1}$ is at most
\begin{equation}
\mathcal{\widetilde{O}}(
\sigma^2k\varepsilon^{-2}
+ \alpha)
\end{equation}
where $\alpha = \max\{ \gamma^2\frac{\sigma_z^2 C_W^2}{\underline{\sigma}^4 R^2}, \gamma^2\frac{C_W^2}{\underline{\sigma}^4}\rho^4, \rho^4\}
\cdot(k + \ln(\frac{4}{\delta}))$.
\end{theorem}

\noindent\textbf{Discussion.} Comparing to the known $\nu$ case in Corollary~\ref{corollary_sample_nu_known}, in this unknown $\nu$ setting we find our algorithm only requires an additional $\varepsilon$-independent number of samples   
$T\beta$ for the sampling complexity from source tasks and $\alpha$ for that from target task to achieve the same performance. \cite{chen2022active} have similar results, but their additional term 
$\beta$ in their Theorem 4.1 
has an order of $\varepsilon^{-1}$. Technically, 
\cite{chen2022active} directly uses the closed form of least square solution and proves that $|\hat{\nu}^2(t)| = \Theta(|\nu^2(t)|), ~\forall t \in [T]$ if $n_t \geq c''\cdot\varepsilon^{-1}$. However, for Lasso-based L1-A-MTRL method, 
we can choose some proper parameter $\lambda_k$ which can upper bound not only the noise term but also the $l_1$-error between Lasso solution and true vector as $\|\hat{\nu}^1-\nu^1\|_1 = \Theta(\|\nu^1\|_1)$ if $n_t \geq c'\cdot\varepsilon^{0}$ (Lemma~\ref{lemma_RE}). Here $c', c''> 0$ are model-related constants.


Moreover, we remark that we have a similar limitation as \cite{chen2022active} that we require some prior knowledge of $\usigma$. However, since they only hit the additional constant terms, they are unlikely to dominate either of the sampling complexities for reasonable values of $d,k,T$ and $\varepsilon \ll 1$. 

Lastly, it is worth mentioning that similar results to Theorem~\ref{thm_noise} also apply when our L1-A-MTRL algorithm incorporates multiple sampling stages, as presented in \cref{alg:L1_A_MTRL_multi} in Appendix. The reason is that we only need to ensure that the minimum sampling budget is larger than $\underline{N}$ which is independent of the stage, and the additional proof follows a similar approach to that of Theorem E.4 in \cite{chen2022active}.

\begin{algorithm}[tb]
   \caption{L1-A-MTRL Method}
   \label{alg:L1_A_MTRL}
\begin{algorithmic}[1]
   \STATE {\bfseries Input:} confidence $\delta$, representation function class $\Phi$, ER bound $\varepsilon \ll 1$, minimum singular value $\underline{\sigma}$
   \STATE Initialize $\underline{N} = \beta_1/T$ with (\ref{equ_beta}) and $\lambda_k$ with (\ref{equ_regular_init}),
   \STATE {\bfseries Phase 1: Exploration} $\nu$  
    \STATE Draw $\underline{N}$ i.i.d samples from every source task datasets
    \STATE Estimate $\hat{\phi}^1, \hat{W}^1$ and $\hat{w}^1_{T+1}$ with Eqn.(\ref{equ_train_on_source}), (\ref{equ_train_on_target})
    \STATE Estimate $\hat{\nu}^{1}$ by
   \textbf{Lasso Program} (\ref{equ_Lasso})
   \STATE Set $\beta_2$ with Eqn. (\ref{equ_beta})
   \STATE {\bfseries Phase 2: Sampling} 
   \STATE Set $n_{t}^2 = \max\{\beta_2 |\hat{\nu}^1(t)|\cdot \|\hat{\nu}^1\|_1^{-1}, \underline{N}\}$.
   \STATE Draw $n_t$ i.i.d samples from $t$-th source task datasets
   \STATE Estimate $\hat{\phi}^2, \hat{W}^2$ and $\hat{w}^2_{T+1}$ with Eqn.(\ref{equ_train_on_source}), (\ref{equ_train_on_target})
\end{algorithmic}
\end{algorithm}

\section{Extentsion: Cost-sensitive Task Selection}
\label{sec_task_selection}

In Section~\ref{sec_main_results}, we proved that the $L_1$ strategy can minimize the total number of samples from the source tasks. 
Implicitly, this assumes the cost of each task is equal, and the first sample costs the same as the $n$-th sample. 
In contrast, we could also consider a non-linear cost function for the $t$-th source task $f_t:\mathbb{N}\rightarrow \R$, which takes in the number of random label query $n$ and outputs the total required cost. 
For example, this could encode the notion that a long-term data subscription from one single source may result in decreasing the average cost over time.

Here we show that, even in this task-cost-sensitive setting, our $L_1$-A-MTRL method \cref{alg:L1_A_MTRL} can still be useful under many benign cost functions. Consider the following example.

\noindent
\textbf{Example: Saltus Cost Function.}
Assume $N_{tot}$ and $\underline{N}$ are fixed. If $n_{t,1} = \underline{N}$, $f :\equiv f_t$ for all $t \in [T]$ and $f$ is composed by fixed cost and linear variable cost:
\begin{equation}\label{fun_saltus_cost}
    f(n) = \left \{
    \begin{array}{cc}
         C_{fix} + C_{var} (n-\underline{N}) &, n > \underline{N}  \\
         0&, n \leq \underline{N}
    \end{array} \right.
\end{equation}
where for each source task $t$ we have $\underline{N}$ free data for sampling.
As a reference, one practical instance for this case is programmatic weak supervision, where setting up a source requires some high cost but afterward, the query cost remains low and linear \citep{zhang2022survey}. If we want to find some proper $\nu$ to minimize the total cost $\sum_{t=1}^Tf_t(n_t)$, then it's equivalent to finding the $L_0$ minimization solution of $\hat{W}\nu=\hat{w}_{T+1}$, where $\hat{W},\hat{w}_{T+1}$ is estimated by free data.  Of course, $L_0$ minimization is known to be intractable, so with proper $\lambda_f$, the $L_1$-A-MTRL method can be a good approximation. 

Now, we are ready to give a formal definition of our goal and the characterization of when our $L_1$-A-MTRL method can be useful. Based on the excess risk upper bound in Theorem~\ref{thm_chen_UB}, to get $ER \leq \epsilon^2$, we are aimed to solve the following optimization problem.
\begin{equation}\label{opt_F_simple}
    \begin{aligned}
\min_{n_{[T],2}}~ &\sum_{t = 1}^T f_t(n_{t,1}+ n_{t,2})\\
        s.t. ~& \sigma^2k(d+T)
        \sum_{t=1}^T \frac{\nu(t)^2}{n_{t,1}+n_{t,2}}
        \lessapprox \varepsilon^2\\
        &W^*\nu = w_{T+1}^*
        \\
        &n_{t,2} \geq 0,  \quad  t\in [T]
    \end{aligned}
\end{equation}
Then we have the following guarantees as long as $f_t$ satisfies the properties shown there.
\begin{theorem}[informal]\label{thm_general_cost_optimize}
Assume $f_t$ is a piecewise second-order differentiable function, and on each sub-function, it satisfies $f_t \geq 0, \nabla f_t \geq 0, \nabla^2 f_t \leq 0$ and $\nabla f_t(n_{t,1}+n_{t,2}) = \Omega( n_{t,2}^{-2+q})$ for some $q \in (0, 2]$.
Denotes the optimal solution of (\ref{opt_F_simple}) as $(n^*_{[T],2}, \nu^*)$. Then under a similar data generation assumption as before, we have
\begin{equation}\label{equ_propto_n_t}
     n_{t,2}^* = h_t(|\nu^*(t)|)
 \end{equation}
where $h_t$ is a monotone increasing function that satisfies: $c_{t,1}x \leq h_t(x) \leq c_{t,2}x^{2/q}$ where $c_{t,1}, c_{t,2} > 0$ . Moreover, we claim A-MTRL algorithm with $n^*_{[T],[2]}$ sampling strategy is $k$-sparse, i.e., $\|n^*_{[T],2}\|_0 \leq k$.
\end{theorem}


\noindent\textbf{Discussion.} If $\nabla f_{t}(n^*_{t,2}) \equiv c > 0$, (\ref{equ_propto_n_t}) is equivalent to $L_1$ strategy mentioned in the previous sections. However, for many other cases, it might be NP-hard to optimize (\ref{opt_F_simple}), such as the Saltus Cost Function example shown above. Therefore, 
our previous algorithm L1-A-MTRL can be widely applied to these task-cost-sensitive scenarios to approximate the optimal strategy.

\begin{figure*}[ht]
\vskip 0.2in
\begin{center}
\begin{subfigure}
	\centering
\includegraphics[width=\columnwidth]{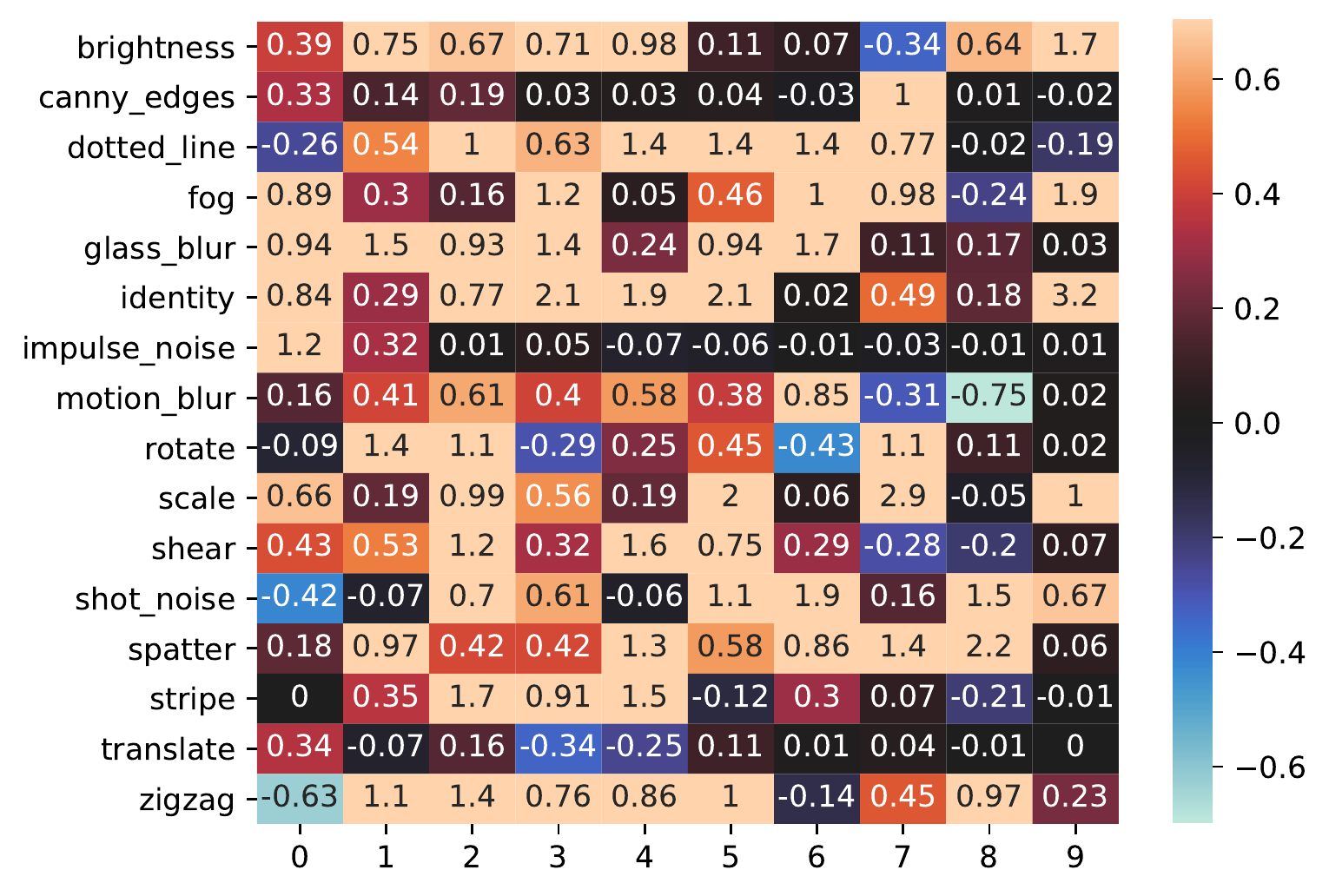}
\end{subfigure}
\begin{subfigure}
	\centering
\includegraphics[width=\columnwidth]{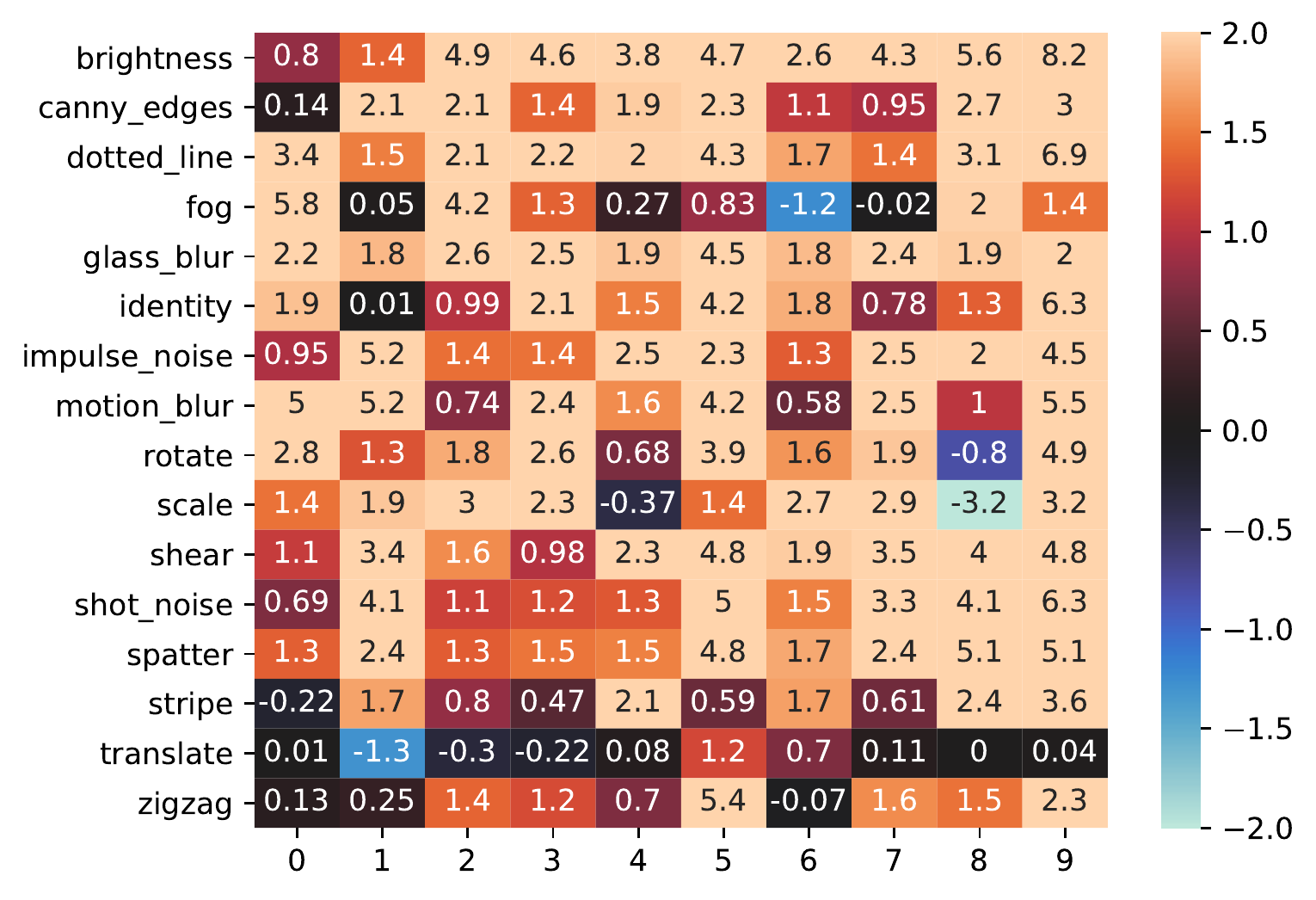}
\end{subfigure}
\caption{\textbf{Performance Comparison.} These pictures show the prediction difference (in \%) between our method and baseline for all target tasks, the larger the better. The y-axis denotes the corruption type while the x-axis denotes to the binarized label, and each grid on $(x, y)$ corresponds to the case that the target task is "\{$y$\}\_\{$x$\}".
\textbf{Left:} \textbf{full tasks scenarios.} Compare L1-A-MTRL and L2-A-MTRL  using linear representation. \textbf{Right:} \textbf{k-task selection scenarios.} Compare two $k$-sparse task selection algorithms L1-A-MTRL and passive-learning  baseline, which randomly selects $k$ source tasks for the second-stage sampling, using Convnet representation.}
\label{pic_full_tasks}
\end{center}
\vskip -0.2in
\end{figure*}

\section{Experiments}
\label{sec_experiment}

Although our theoretical analysis only holds for a linear representation,
our  experiments also show the effectiveness of our algorithm on neural network representations as well in the task selection case. 
In this section, we follow the experimental settings in \cite{chen2022active} and empirically evaluate L1-A-MTRL on the corrupted MNIST (MNIST-C) dataset proposed in \cite{mu2019mnist}. We reflect the preponderance of our algorithm on the two scenarios mentioned above. The first one is cost-agnostic, which aims to minimize the total sampling number from the source tasks and can reach all the source tasks.
Another scenario is task-cost-sensitive like Section~\ref{sec_task_selection} and we 
 particularly concentrate on $k$ task-selection algorithms which correspond to cost functions 
 like saltus cost function, and the learner is only allowed to sample from only $k$ tasks after the initial exploration stage.
We call the first case \textbf{full task} scenario and the second one \textbf{k-task selection} scenario for convenience.
Please refer to Appendix~\ref{sec_app_k_task_illu} for further illustration of our intuition for the k-task selection scenario.


\subsection{Experimental Setup}
\noindent
\textbf{Datasets.} The MNIST-C dataset is a comprehensive suite of 16 corruptions applied to the MNIST test set. Like in \cite{chen2022active}, we divide each corruption-related sub-dataset into 10 tasks according to their labels ranging from $0\sim 9$ and thus get 160 separate new tasks denoted by "\{corruption type\}\_\{label\}". For instance, \textit{brightness\_0} denotes the data  corrupted by brightness noise and relabeled to $1/0$ based on whether the data corresponds to number 0 or not. And once we choose 1 task called "\{type A\}\_\{label B\}" for the target task, the other 150 tasks that don't contain "type A" corruption will be chosen as source tasks.

\noindent
\textbf{Experimental Setups and Comparisons.}  Like in \cite{chen2022active}, we replace the cross-entropy loss, which is commonly used for MNIST, with the regression $l_2$ loss in order to align with the theoretical setting in this paper.
As the model setting, for full tasks scenario, we use the linear representation as defined in our theorem. We set $d = 28*28$, $k = 50$ and there are $T=150$ source tasks in total. And we compare 
L1-A-MTRL and L2-A-MTRL\cite{chen2022active} algorithms on the above datasets with 160 different target tasks.
For the k-task selection scenario, we use a 2-layer ReLU CNN followed by a fully-connected linear layer as the representation map. Since neural networks can better capture the feature, here we set a smaller representation dimension $k=10$ to show the advantage of the sparse task selection algorithm while other parameters follow the setting in the case of the full tasks.
We compare L1-A-MTRL, which has been proved to be $k$-sparse from Theorem~\ref{thm_worst_target_ER}, together with vanilla $k$-sparse baseline that randomly selects $k=10$ source tasks for sampling data at the second stage. Please refer to Appendix~\ref{appendix_exp_algo} for details of algorithm implementation and Appendix~\ref{appendix_choose_lambdak} for  details on how to determine the value of $\lambda_k$.

\subsection{Results}
\textbf{Full tasks scenario.} 
In summary, L1-A-MTRL achieves $0.54\%$ higher average accuracy among all the target tasks than L2-A-MTRL and results same or better performance in 126 out of 160 tasks. Due to the imbalanced dataset, $10\%$ is the error rate of the baseline which randomly guesses the label, and the average prediction incorrect rate for L2-A-MTRL is $7.4\%$. 

\textbf{k-Task selection scenario.} Similarly, L1-A-MTRL achieves $2.2\%$ higher average accuracy among all the target tasks than the vanilla baseline which has the average prediction error rate of $5.7\%$. And our algorithm results in the same or better performance in 149 out of 160 tasks. This shows the effectiveness of our method on neural network representation.

In Section~\ref{sec_add_exp_N} of the appendix, we provide additional comparisons of the empirical sampling budgets for different algorithms. The results demonstrate that L1-A-MTRL requires fewer samples compared to L2-A-MTRL and P-MTRL while achieving comparable performance. These findings further underscore the effectiveness of our L1-A-MTRL algorithm.

\section{Conclusion}

We introduced a novel active sampling strategy L1-A-MTRL to sparse sample from target-related source tasks and learn a good representation that helps the target task.
From a theoretical perspective,  we first showed that L1-A-MTRL is strictly better than the previous L2-A-MTRL by proving a novel sampling-strategy-dependent lower bound and then provided a tighter upper bound correspondingly. From the empirical perspective, we showed our algorithm is not only effective under the standard setting but can achieve even better results in the practical scenario where the number of source tasks is restricted.

\section*{Acknowledgements}
SSD acknowledges the support of NSF IIS 2110170, NSF DMS 2134106, NSF CCF 2212261, NSF IIS 2143493, NSF CCF 2019844, NSF IIS 2229881.



\bibliography{reference}
\bibliographystyle{icml2023}

\newpage
\setcounter{tocdepth}{2}
\appendix
\onecolumn



\section{Related Work}\label{appendix_related_work}

\textbf{Empirical works on P-MTRL and A-MTRL.}
Multi-task representation learning has been widely applied and achieved great success in the natural language domain  GPT-2 \citep{radford2019language}, GPT-3\citep{brown2020language}, vision domain CLIP \citep{radford2019language} and multi-model Flamingo \citep{alayracflamingo}. Nevertheless, such large models are costly in both data collecting/cleaning and training. Recently, many works focus on efficiently selecting the source task. In the natural language domain, for example, \cite{yao2022nlp} use a heuristic retriever method to select a subset of target-related NLP source tasks; More recently, works like \cite{asai2022attentional,zhang2022task} use prefix/prompt to capture the relation between source and target tasks. Similar topics have also been studied in the vision domain, for example, \cite{zamir2018taskonomy} propose a transfer learning algorithm based on learning the underlying structure among visual tasks, which they called Taskonomy, and there are many following works propose different approaches on this Taxonomy dataset, including \cite{fifty2021efficiently, standley2020tasks}.


\textbf{Theoretical works on P-MTRL.} 
There are also many existing works on provable P-MTRL. \citet{tripuraneni2020theory, Tripuraneni2021ProvableMO, Du2021FewShotLV,thekumparampil2021sample,collins2021exploiting,xu2021representation} assume there exists a ground truth shared representation across all tasks. In particular, \citet{tripuraneni2020theory, Tripuraneni2021ProvableMO, thekumparampil2021sample} assume a low-dimension linear representation like us while \citet{Du2021FewShotLV} generalize to both high-dimensional representation and 2-layer Relu network. \citet{tripuraneni2020theory} also further considers any general representation with linear predictors. Both works obtain similar results. Besides, many recent works focus on fine-tuning in theoretical contexts \cite{shachaf2021theoretical,chua2021fine,chensx2021weighted,kumar2022fine}. 

For the lower bound, for the first time,  \citet{Tripuraneni2021ProvableMO} proves a minimax lower bound for the estimation error of the estimated representation layer measured by subspace angle distance. But we claim it can't directly deduce a similar lower bound of the test error on the target task, which relates to one of our main contributions. 
The reason is that though the estimated representation may be far away from the ground truth one, the learner can estimate a proper target predictor to achieve a sufficiently small test error as long as $B^*w_{T+1}^*$ (almost) lies in the column space of $\hat{B}$, where the notations are defined in the preliminary.


\textbf{Theoretical works on A-MTRL.} 
In order to overcome the problems in P-MTRL, some subsequent works focused on giving different priorities to the source tasks by methods like active learning \cite{chen2022active} and weighted training \cite{chensx2021weighted}. Representatively, \citet{chen2022active} is the first work to propose A-MTRL which calculates the proper sampling number for each source task. It iteratively estimates the relevance of each source task to the target task by estimating the relevance vector $\nu^*$. \citet{chen2022active} utilizes the $L_2$ strategy defined in Def.~\ref{def_LpNq} to decide the sampling strategy and significantly outperforms passive MTRL (P-MTRL), which uniformly samples from the source tasks, both theoretically and empirically. Nevertheless, the optimal sample strategy for A-MTRL is underexplored, and the non-sparsity of $\nu^2$ may cause inconvenience for task-cost-sensitive scenarios. We develop our works based on the problem setting in \cite{chen2022active} and propose a more efficient sampling strategy.
As another approach, \citet{chensx2021weighted} concentrates on learning a weighting over the tasks. The crucial difference between
their work with ours is that they can attach to the whole dataset whereas we assume not but actively query new data from some large datasets (e.g., the task represented by the search terms to Wikipedia or Google). They also assume that some tasks may not only be irrelevant but even harmful and need to be down-weighted. 


\section{Technical Notations}\label{appendix_notations}
We summarize the technical notations used in the appendix as follows.


\noindent
\textbf{Grassmann Manifold.} Assume $d \geq k$, we denote by $Gr_{d, k}$ the Grassmann manifold which contains all the subspaces that are spanned by $k$ linearly independent $d$-dimensional vectors. For $d \geq k$, we let $O_{d,k}$ be the set of matrices whose column contains $k$ orthonormal vectors that are in $\R^{d}$. Then any $B \in O_{d, k}$ corresponds to an element, which is spanned by the column vectors of $B$, of $Gr_{d, k}$. Actually, an element in $Gr_{d, k}$ is corresponds to an equivalent class of $d\times k$ matrices that satisfies the equivalent relation $\sim$:
\begin{equation}
    Y \sim X \Leftrightarrow Y = XA, \ \forall A \in GL(k, \R)
\end{equation}
where $GL(k, \R)$ denotes general linear group over $\R$ of degree $k$.

\noindent
\textbf{Subspace Distance.} Finally, we use the same definition as \cite{Tripuraneni2021ProvableMO} and \cite{Pajor1998MetricEO} to define the distance between the subspaces in the Grassmann manifold. We let $s_p(T) = (\sum_{i\geq 1} |\sigma_i(T) |^p)^{1/p}$ for any matrix $T$ and any $p \in [1, \infty]$. In particular, $s_{\infty}$ is the operator norm of $T$. For $E, F \in O_{d, k}$, from Proposition 6 of \cite{Pajor1998MetricEO} we define 
$s_q(E,F) = (2\sum_{i= 1}^{k} |1 - \sigma_i^2(E^TF) |^{q/2})^{1/q}$ 
to be the subspace distance between the spaces spanned by the column vectors of $E$ and $F$, respectively. Particularly,  $s_{\infty}(E,F) = \sqrt{1 - \sigma_k^2(E^TF)}$.

\section{Proof of Theorem~\ref{thm_optimal_sample_general}}
\label{appendix_sec_best_n_T}
\noindent\textbf{Proof of Lemma~\ref{lemma_nu}}. 
We can use the following equivalent optimization problem to prove our Lemma:
\begin{equation}\label{opt_nu_tilde}
\begin{aligned}
    &\min_{n_{[T]}}& G(n_{[T]})&:=\sum_{t=1}^T \frac{|\nu^*(t)|^2}{n_t}\\
    &s.t.& c_0(n_{[T]})&:= N_{tot} - \sum_{t=1}^Tn_t = 0\\
    &&c_{t}(n_t)&:= n_t - \underline{N} > 0, 
    \quad \forall t \in [T]
\end{aligned}
\end{equation}
The corresponding Lagrangian function for (\ref{opt_nu_tilde}) is
\begin{equation}
    L(n_{[T]}):= G(n_{[T]}) - \lambda_0 c_0(n_{[T]}) - \sum_{t=1}^T \lambda_t c_t(n_t)
\end{equation}
Then from the Karush-Kuhn-Tucker condition, for all $t \in [T]$ we have the necessary condition
\begin{equation}
    \begin{aligned}
        \frac{\partial L}{\partial n_t} &= -\frac{|\nu^*(t)|^2}{n_t^2} + \lambda_0 - \lambda_t = 0\\
        \lambda_t &\geq 0\\
        \lambda_t c_{t}(n_t) &= \lambda_t(n_t - \underline{N})=0
    \end{aligned}
\end{equation}

So we get $\lambda_0 > \lambda_t \geq 0,~ \forall t \in [T]$ and 
\begin{equation}
        n_t = \left\{
    \begin{array}{cc}
    \lambda_0^{-0.5}|\nu^*(t)|&, \lambda_t = 0 \Rightarrow n_t \geq \underline{N},\\
    &\\
    \underline{N}  &, \lambda_t > 0 \Rightarrow n_t = \underline{N}.
    \end{array}
    \right.
\end{equation}

thus we finish the proof.

As a supplement, we give another proof for the special case in this Lemma where we assume $n_t > \underline{N}$ for every $t \in [T]$.
Let $\beta(t) := \frac{\nu^*(t)}{\Vert \nu^*\Vert_2}$, $\alpha_t = \frac{n_t}{N_{tot}}$ and thus $\sum_{t=1}^T\beta^2(t) =\sum_{t=1}^T\alpha_t = 1$. Therefore by Cauchy inequality, 
\begin{equation}\label{inequ_proof_cauchy}
\begin{split}
    \Vert\tnu^*\Vert ^ 2_2 
    &= \frac{\Vert \nu^*\Vert_2^2}{N_{tot}}\sum_{t=1}^T\frac{\beta^2(t)}{\alpha_t}\\
    &= \frac{\Vert \nu^*\Vert_2^2}{N_{tot}}(\sum_{t=1}^T\frac{\beta^2(t)}{\alpha_t})(\sum_{t=1}^T\alpha_t)\\
    &\geq \frac{\Vert \nu^*\Vert_2^2}{N_{tot}}(\sum_{t=1}^T|\beta(t)|)^2=\frac{\Vert\nu^*\Vert_1^2}{N_{tot}}
\end{split}
\end{equation}

The equality in (\ref{inequ_proof_cauchy}) is achieved iff $\frac{|\beta(t)|}{\sqrt{\alpha_t}} = c \sqrt{\alpha_t}$ for evert $t \in [T]$ with $c > 0$, which means that $n_t$ is proportional to $|\nu^*(t)|$.
\qed

\noindent\textbf{Proof of Corollary~\ref{corollary_sample_nu_known}.} 
As stated in Lemma~\ref{lemma_nu}, $n_t^* = \max\{c^\prime|\nu(t)|, \underline{N}\}$ and $c^\prime > 0$ is some constant such that $\sum_{t=1}^Tn_t^* = N_{tot}$, so we have $c^\prime|\nu(t)| \leq n_t^* \leq c^\prime|\nu(t)| + \underline{N}$. Sum up both sides of the inequality for all $t \in [T]$, then:
\begin{equation}\label{eq:c_and_N_tot}
    c^\prime \Vert \nu\Vert_1 \leq N_{tot} \leq c^\prime \Vert \nu\Vert_1 + T \underline{N}
\end{equation}
Therefore, if we assume $N_{tot} \gg T \underline{N}$, then we get $N_{tot} = (1 + o(1)) c^\prime \Vert\nu\Vert_1$. 
In fact, we only need to ensure that $N_{tot} > 2T\underline{N}$, which results in $c' > N_{tot}/(2|\nu|_1)$,  since the coefficient in the error bound (\ref{inequ_ER_upper_wildenu}) is unconsidered.

Let $S_1 = \{t \in [T]| n_t \geq \underline{N}, |\nu(t)| > 0\}$ and $S_2 = \{t \in [T]| n_t < \underline{N}, |\nu(t)| > 0\}$. Then for any fixed $\nu$, from Lemma~\ref{lemma_nu}, we have the following inequality for the optimal strategy:
\begin{equation}\label{inequ_tilde_nu_UB}
    \|\widetilde{\nu}\|_2^2 
    = \sum_{t \in S_1} \frac{\nu(t)^2}{n_t} + \sum_{t \in S_2} \frac{\nu(t)^2}{\underline{N}} 
    \leq \sum_{t \in S_1 \cup S_2} \frac{\nu(t)^2}{c'|\nu(t)|} 
    =(1+o(1))\sum_{t \in S_1 \cup S_2} \frac{|\nu(t)|}{N_{tot}}\|\nu\|_1 
    = (1+o(1))\frac{\|\nu\|_1^2}{N_{tot}}
\end{equation}
Here the inequality holds if and only if
$S_2$ is empty, which means that for all $t \in [T]$, $\nu(t)$ must satisfies $\nu(t) = 0$ or
$c'|\nu(t)| \geq \underline{N}$.
Combining (\ref{inequ_tilde_nu_UB}) and Theorem~\ref{thm_chen_UB}, we get the results.
\qed

\noindent\textbf{Proof of Theorem~\ref{thm_optimal_sample_general}.}
From (\ref{inequ_tilde_nu_UB}) we know that if $c'|\nu(t)| \geq \underline{N}$ for all $t \in [T]$ such that $|\nu(t)| > 0$
, then $n_{t}^* = N_{tot}|\nu(t)|/{\Vert \nu\Vert_1}, ~\forall t \in \{t' \in [T]||\nu(t')| > 0\}$, and $\Vert\tnu\Vert ^ 2_2$ attain its minimum $\Vert\nu\Vert_1^2/{N_{tot}}$.

We prove that such a condition can be achieved when $N_{tot}$ is sufficiently large.
Assume that $c’|\nu(t)| < \underline{N}$ always holds for some $t \in [T]$ where $\nu(t) \neq 0$. Then if we choose $N_{tot} = T\underline{N} + \underline{N}/{|\nu(t)|}\Vert\nu \Vert_1$, we will have $c’|\nu(t)| = c’ \Vert\nu\Vert_1 \cdot |\nu(t)|/{\Vert\nu\Vert_1} \geq (N_{tot}-T\underline{N})|\nu(t)|/{\Vert\nu\Vert_1} \geq \underline{N}$, where we use the fact that $\underline{N}T + c’\Vert \nu\Vert_1 \geq N_{tot}$ from Eqn.~\ref{eq:c_and_N_tot}. This is contradicted by the assumption, and thus we can always find some $N_{tot}$ such that $c’|\nu(t)| \geq \underline{N}$ if $\nu(t) \neq 0$.

So for any given $\nu$, the optimal sampling strategy $n_{t}(\nu)$(Lemma~\ref{lemma_nu}) can let $\| \widetilde{\nu}\|_2^2$ achieves its minimum $\|\nu\|_1^2/{N_{tot}}$. Then we vary $\nu$ among the solution candidate set of $W^*\nu = w_{T+1}^*$ and find $L_1$-minimization solution $\nu^1$ can minimize $\|\nu\|_1^2/{N_{tot}}$. Therefore, $(\nu^1, n_{[T]}^1)$ is optimal for the original problem (\ref{opt_min_nu_general}).
\qed

\section{Proof of Theorem~\ref{thm_worst_target_ER}}\label{appendix_sec_worst}
\subsection{Preparations for minimax lower bound}
First, we reclaim some concentration inequalities commonly used in the previous work~\cite{Du2021FewShotLV, chen2022active}.
\begin{lemma}\label{lemma_sigma_concen}
(A variant of Lemma A.6 in \cite{Du2021FewShotLV}) Let $a_1,...,a_n$ be i.i.d. $d$-dimensional random vectors such that $\mathbb{E}[a_i] = 0$, $\mathbb{E}[a_ia_i^\top] = I$, and $a_i$ is $\rho^2$-subgaussian. For $\delta \in (0,1), \epsilon \in (0,\frac{1}{2})$, suppose $n > \frac{1 }{\epsilon^2}c_a\rho^4(d + \ln(\frac{1}{\delta}))$ for some universal constant $c_a$. Then with probability at least $1-\delta$ we have
\begin{equation}
    (1-2\epsilon) I_d \preceq \frac{1}{n}\sum_{i=1}^n a_i a_i^\top \preceq (1+2\epsilon) I_d
\end{equation}
\end{lemma}
\qed

Recall that $\Sigma_t^* = \E_{x_t\sim p_t}[x_t x_t^\top]$ and $\hat{\Sigma}_{t} := \frac{1}{n_t}(X_t)^\top X_t$ for any $t \in [T+1]$, then we have:
\begin{lemma}\label{lemma_sigma_kd}
(A variant of Claim A.1, A.2 in \cite{Du2021FewShotLV})
Suppose for $\delta \in (0,1)$. Let $n_t > \frac{1 }{\epsilon^2}c_a\rho^4(d + \ln(\frac{2T}{\delta}))$ for all $t\in[T]$, then with probability at least $1-\frac{\delta}{2}$ over the inputs $X_1, ... ,X_T$ in the source tasks, we have
\begin{equation}\label{inequ_Sigma}
    (1-2\epsilon) \Sigma_t \preceq 
    \hat{\Sigma}_{t}
    \preceq (1+2\epsilon) \Sigma_t
\end{equation}
Here $c_a>0$ is a universal constant. 
Similarly, let $n_{T+1} > \frac{1 }{\epsilon^2}c_a\rho^4(k + \ln(\frac{2}{\delta}))$. Then for any given matrix $B_1, B_2\in\R^{d\times k}$ that is independent of $X_{T+1}$ , with probability $1-\frac{\delta}{2}$  over $X_{T+1}$ we have
\begin{equation}\label{inequ_BSigmaB}
    (1-2\epsilon) B_1^\top \Sigma_{T+1} B_2
    \preceq 
    B_1^\top\hat{\Sigma}_{T+1}B_2
    \preceq (1+2\epsilon) B_1^\top \Sigma_{T+1} B_2
\end{equation}
\end{lemma}
\qed

And then, we show that $\sum_{t=1}^T|X_t(\hat{B}\hat{w}_t - B^*w_t^*)|^2 \asymp \sigma^2(kT + k(d-k))$. The upper bound has been shown in Claim A.3 in \cite{Du2021FewShotLV}, and the lower bound will be shown in the following theorem.
\begin{theorem}\label{thm_XBW_LB}
With conditions in Theorem~\ref{thm_worst_target_ER}, with probability $1-\delta$ we have:
\begin{equation}\label{inequ_XBW_LB_minimax}
    \inf_{(\hB,\hW)} \sup_{(\Bs,\Ws)}\sum_{t=1}^T|X_t(\hat{B}\hat{w}_t - B^*w_t^*)|^2 \gtrsim \sigma^2(kT + k(d-k))
\end{equation}
The key theorems and lemmas are as follows.
\end{theorem}
\begin{theorem}\label{thm_G1}
Let $G_0 := \{BW | B \in O_{d,k} \ ; W \in \R^{k\times T} \}$, and $G_1(\delta_1) := \{BW | B \in O_{d,k} \ ; W \in \R^{k\times T} \ ; \Vert W \Vert_F \leq \delta_1 
, t \in [T]\}$ be a local packing of $G_0$, where $w_t$ is the $t$-th column vector of $W$. Then there is a lower bound for $G_1$'s packing number:
\begin{equation}
\ln M(G_1(\delta_1), \Vert \cdot \Vert_F, \Delta_1) \gtrsim k(d-k) + kT
\end{equation}
where $\Delta_1$ will be determined soon.
\end{theorem}

\begin{lemma}\label{lemma_Gr}
[Adapted from \cite{Pajor1998MetricEO}] For any $1 \leq k \leq d$ such that $k \leq d - k$, for every $\epsilon > 0$, we have
\begin{equation}
    (\frac{c_1}{\epsilon})^{k(d - k)} \leq N(Gr_{d,k}, s_{\infty}, \epsilon) \leq (\frac{c_2}{\epsilon})^{k(d - k)}
\end{equation}
with universal constants $c_1, c_2 > 0$. From the relation between packing number and covering number~\cite{wainwright_2019}, we have:
\begin{equation}
    M(Gr_{d,k}, s_{\infty}, \epsilon) \geq (\frac{c_1}{\epsilon})^{k(d - k)} 
\end{equation}
\hfill\qedsymbol
\end{lemma}

\begin{lemma}\label{lemma_BW}
Let $B^1, B^2 \in O_{d,k}$, $w^1, w^2 \in \R^{k}$. With SVD we get $(B^1)^\top B^2 = P D Q^T$, where $P, Q \in O_{k,k}$, $D = diag(\sigma_1, ..., \sigma_k)$. Obviously $\sigma_i \in [0,1]$, and we define $v^1 = P^\top w^1$, $v^2 = Q^\top w^2$. If subscripts denotes the index of vectors, we have:
\begin{equation}\label{equ_BW_eq}
    |B^1w^1 - B^2w^2|^2 = \sum_{i=1}^k[2|v_i^1||v_i^2|f(v_i^1, v_i^2) + (|v_i^1|-|v_i^2|)^2]
\end{equation}
where 
\vspace{-2em}
\begin{center}
\begin{equation}\label{equ_f}
f(v_i^1, v_i^2)=\left\{
\begin{aligned}
1 - \sigma_i, \qquad & sign(v_i^1\cdot v_i^2) = 1\\
1 + \sigma_i, \qquad  & sign(v_i^1\cdot v_i^2) = -1 
\end{aligned}
\right.
\end{equation}
\end{center}
And we can get the lower bound:
\begin{equation}\label{equ_BW_lower}
    |B^1w^1 - B^2w^2|^2 \geq 2|v_k^1||v_k^2|(1 - \sigma_k) + \sum_{i=1}^{k}(|v_i^1|-|v_i^2|)^2 \geq 0
\end{equation}
\end{lemma}

\noindent\textbf{Proof of  Lemma~\ref{lemma_BW}.} By the calculation we get this result:
\begin{equation}
\begin{split}
    |B^1w^1 - B^2w^2|^2 &= (B^1w^1 - B^2w^2)^\top(B^1w^1 - B^2w^2)\\
    &= |w^1|^2 + |w^2|^2 - 2(w^1)^\top (B_1)^\top B_2 w^2\\
    &= |v^1|^2 + |v^2|^2 - 2(v^1)^\top D v^2\\
    &= \sum_{i=1}^k((v^1_i)^2 + (v^2_i)^2 - 2 v^1_i v^2_i \sigma_i)
\end{split}
\end{equation}
To make each term of the equation above non-negative, we use sign function:
\begin{equation}
\begin{split}
    |B^1w^1 - B^2w^2| ^2
    &= \sum_{i=1}^k[(v^1_i)^2 + (v^2_i)^2 - 2 sign(v^1_i v^2_i)\times v^1_i v^2_i + 2 v^1_i v^2_i (sign(v^1_i v^2_i)-\sigma_i)]\\
    &= \sum_{i=1}^k[(v^1_i)^2 + (v^2_i)^2 - 2 |v^1_i| |v^2_i| + 2 |v^1_i| |v^2_i| (1-sign(v^1_i v^2_i)\sigma_i)]\\
    &= \sum_{i=1}^k[(|v_i^1|-|v_i^2|)^2 + 2|v_i^1||v_i^2|f(v_i^1, v_i^2) ]
\end{split}
\end{equation}
\qed

Besides, we begin to construct a separate set for $G_1$. Firstly we let $G_B = \{B^1, ..., B^{M_B}\}$ be a $\epsilon_{B}$-separated set for metric $s_{\infty}$ in $Gr_{d,k}$, where $\epsilon_{B} \leq \min(\frac{c_1}{2}, 1)$ as $c_1$ in Lemma~\ref{lemma_Gr}. 

Then denote $(B^m)^\top B^n = P(m,n)D(m,n)Q(m,n)$, where $P(m,n), Q(m,n) \in O_{k,k}$, $D(m,n) = diag(\sigma_1(m,n), ..., \sigma_k(m,n))$, and $P(m,n) = Q(m,n) = D(m,n) = I_k$ iff $m=n$. On the other hand, for $t \in [T]$, we denote $v^j_{t,i}(P(m,n))$ to be the $i$-th component of $v^j_{t}(P(m,n)) := P(m,n)^\top w^j_t$, and similarly for $v^j_{t}(Q(m,n)) := Q(m,n)^\top w^j_t$.

\begin{lemma}\label{lemma_pack}
Suppose $G_{V} = \{V^j = (v_1^j,...,v_T^j) | j \in S, v_t^j \in \R^{k}, v_t^j$ satisfy Equ.~\ref{equ_v_cond} and attain largest $ |S| \}$:
\vspace{-2em}
\begin{center}
\begin{equation}\label{equ_v_cond}
\begin{aligned}
&|v_{t,k}^j| \geq \frac{\delta_{V}}{\sqrt{T}\epsilon_{B}}, & \forall j, \ \forall t \in [T] \\
&\Vert V^j\Vert_F = \sum_{t=1}^T|v_{t}^j|^2 \leq \frac{C_V\delta_{V}}{\epsilon_{B}}, & \forall j \\
&\Vert V^i - V^j\Vert_F = \sum_{t=1}^T|v_{t}^i - v_{t}^j|^2 \geq \frac{\delta_{V}}{\epsilon_{B}}, & \forall i, j \\
\end{aligned}
\end{equation}
\end{center}
\vspace{-0.5em}
where $C_V$ is a universal constant and $4 < C_V < 5$. For $m,n \in [M_{B}]$, let $G_W(P(m,n)) := \{ W^j = (w_1^j, ..., w_T^j) \ | \ \exists \ V^j \in G_{V} , s.t. \ W^j = P(m,n)V^j  \}$ and similarly for $G_W(Q(m,n))$. Then let $G_{BW} = \{(B,W) |  \ \exists \ m,n \in [M_{B}], \ W^m \in G_W(P(m,n)), \ W^n \in G_W(Q(m,n)),  s.t. \ BW \in \{B^mW^m, \ B^n W^n\}\}$, and we claim that $G_{BW}$ is a $\delta_V$-separated subset of $G_1$ with Frobenius norm.
\end{lemma} 

\noindent\textbf{Proof of  Lemma~\ref{lemma_pack}.} 
For each $t \in [T]$, we divide into 2 cases:

\textbf{Case 1.} For the case $m \neq n$, we will work out the lower bound of Equ.~\ref{equ_BW_lower}. Since for any $m \neq n$:
\begin{equation}
\begin{split}
    &s_{\infty}(m,n) = \sqrt{1-\sigma_k^2((B^m)^\top B^n)} \geq \epsilon_B^2\\
    \Rightarrow &1-\sigma_k((B^m)^\top B^n) \geq \frac{\epsilon_B^2}{1+\sigma_k((B^m)^\top B^n)}> \frac{\epsilon_B^2}{2}
\end{split} 
\end{equation}

combined with the first inequality of Equ.~\ref{equ_v_cond}, we know by the definition of $G_{Bw}$, there exist some $i, j$ such that:
\begin{equation}
    \begin{split}
        \sum_{t=1}^T|B^mw_t^m - B^nw_t^n|^2 \geq 2\sum_{t=1}^T|v_{t,k}^i||v_{t,k}^j|(1 - \sigma_k) \geq \delta_{V}^2
    \end{split}
\end{equation}

\textbf{Case 2.} For the case $m = n$, note that $\sigma_i = 1$ for all $i \in [k]$. Combined Equ.~\ref{equ_BW_eq} , Equ.~\ref{equ_f}, Equ.~\ref{equ_v_cond} and condition $\epsilon_{B} < \min(\frac{c_1}{2}, 1)$, there exist some $i, j $ such that:
\begin{equation}
    \begin{split}
    \sum_{t=1}^T|B^mw_t^m - B^mw_t^m|^2 = \sum_{t=1}^T\sum_{l=1}^{k}(v_{t,l}^i-v_{t,l}^j)^2
    = \sum_{t=1}^T|v_{t}^i-v_{t}^j|^2
    \geq \frac{\delta_V^2}{\epsilon_B^2}\geq \delta_V^2
    \end{split}
\end{equation}
Combined them together, we see that for any $m, n \in [M_B]$, any $ W^m \in G_W(P(m,n)), \ W^n \in G_W(Q(m,n))$ such that $B^m = B^n, W^m = W^n$ not hold in the meantime, we have:
\begin{equation}
    \Vert B^mW^m - B^nW^n\Vert_F = \sum_{t=1}^T |B^mw_t^m - B^mw_t^m|^2 \geq \delta_V
\end{equation}
\qed

\noindent\textbf{Proof of  Theorem~\ref{thm_G1}.} 

From the construction  in Lemma~\ref{lemma_pack},  we consider flattening $V^j$ into a $k\times T$ vector $\eta^j \in \R^{kT}$, where $V^j \in G_{V} = \{V^j = (v_1^j,...,v_T^j) | j \in S, v_t^j \in \R^{k}, v_t^j$ satisfy Equ.~\ref{equ_v_cond} and attain largest $ |S| \}$. Then the last two conditions in (\ref{equ_v_cond}) show that $\eta^j$ is a $\frac{\delta_V}{\epsilon_B}$-separated set contained in a ball of
radius $\frac{C_V\delta_V}{\epsilon_B}$ in $l_2$-norm. Actually, the first condition means removing the small central part along very axis of $\eta^j$ in the above ball, and it's clear to see that $G_V$ has the same order of the cardinality if we drop the first inequality of (\ref{equ_v_cond}). So if we use $card$ to denote the cardinality of a set, we get:

\begin{equation}\label{inequ_card_G_V}
    \ln(card(G_{V})) \gtrsim kT 
\end{equation}

Then from the definition of $G_W$ and $G_{BW}$ in Lemma~\ref{lemma_pack}, we see that:
\begin{equation}
    \begin{split}
    \ln(card(G_{BW})) &= \ln((\frac{M_B(M_B-1)}{2}\times 2 + M_B)\cdot \ln(card(G_W)))\\
    &= 2\ln(M_B) +\ln( card(G_{V}))\\
    &\gtrsim k(d-k)\ln(c_1/\epsilon_B) + kT, \qquad(\ref{inequ_card_G_V})\\
    &\gtrsim k(d-k)+kT, \qquad(\epsilon_{B} < \min(\frac{c_1}{2}, 1))\\
    \end{split}
\end{equation}
Choose $\Delta_1 = \delta_1\epsilon_B/C_V$
and we finish the proof.
\qed

\noindent\textbf{Proof of  Theorem~\ref{thm_XBW_LB}.}

Note that $\underline{\lambda} = \sigma_{\min}(\Sigma_t^{1/2})$,  $\overline{\lambda} = \sigma_{\max}(\Sigma_t^{1/2})$ and $\kappa = \overline{\lambda}/\underline{\lambda}$, we can construct the local packing following Lemma~\ref{lemma_pack} by using $\widetilde{W}$ to replace $W$ where $\widetilde{w}_t = \sqrt{n_t}w_t$. And we choose $\delta_1^\prime = 0.9 \delta_1$ where $\delta_1 = \frac{\delta_V}{\epsilon_B}$. Then we have:
\begin{equation}\label{inequ_sum_XBw}
\begin{split}
    \sqrt{\sum_{t=1}^T\Vert X_t(B^iw^i_t - B^jw^j_t)\Vert_2^2}
    &\leq 1.1\overline{\lambda} \Vert B^i\widetilde{W}^i - B^j\widetilde{W}^j\Vert_F\\
    &\leq 1.1\overline{\lambda}\cdot C_V\delta_{1}\cdot \frac{\delta_1^\prime}{0.9\delta_1} \\
    &< 6\overline{\lambda}\delta_{1}^\prime\\
\end{split}
\end{equation}
\begin{equation}\label{inequ_sum_XBw_lb}
\sqrt{\sum_{t=1}^T\Vert X_t(B^iw^i_t - B^jw^j_t)\Vert_2^2} \geq 0.9 \underline{\lambda}\Vert B^i\widetilde{W}^i - B^j\widetilde{W}^j\Vert_F \geq \delta_{1}^\prime \underline{\lambda}
\end{equation}

Here for convenience we choose $C_V = 4.5$, and this will just influence the universal constant since $C_V$ is $\Theta(1)$ as in Lemma~\ref{lemma_pack}.
Note the sum of excess risks on the source tasks in (\ref{inequ_sum_XBw}), (\ref{inequ_sum_XBw_lb}) is actually a semi-metric between $(B^{i}, W^{i})$ and $(B^{j}, W^{j})$, and it's easy to construct the corresponding $\delta_1^\prime \underline{\lambda}$-separated 
set $G_{BW}$ from $G_{B\widetilde{W}}$ set obtained in 
Lemma~\ref{lemma_pack}. We recall that $Y_t = X_tB^*w_t^* + Z_t$, and define $Y_t \sim \mathbb{P}_{t}^{\,j}$ where $\mathbb{P}_{t}^{\,j} = \mathcal{N}(X_tB^*w_t^*, \sigma^2 \mathbb{I}_{n_t})$. And we further let $\mathbb{P}^{\,j}:= \prod_{t=1}^T\mathbb{P}_{t}^{\,j}$. Then by the independency among every tasks, we have the Kullaback-Leibler divergence:
\begin{equation}
\begin{split}
    D(\mathbb{P}^{\,i}\,\Vert\  \mathbb{P}^{\,j}) &= \sum_{t=1}^TD(\mathbb{P}_t^{\,i}\, \Vert\ \mathbb{P}_t^{\,j})\\
     &= \frac{1}{2\sigma^2}\sum_{t=1}^T\Vert X_t(B^iw_t^i - B^jw_t^j)\Vert_2^2\\
     &\leq \frac{18\overline{\lambda}^2(\delta_{1}^\prime)^2}{\sigma^2} \qquad(\ref{inequ_sum_XBw})
\end{split}
\end{equation}

Note that $G_{BW}$ is a $\delta_1^\prime \underline{\lambda}$-separated set over $G_1$, which is a local packing of $G_0$, we then let $M = M(G_0,\Vert\cdot\Vert_F ,(\delta_1^\prime)^2)$ and have the following Fano's lower bound \cite{wainwright_2019}:
\begin{equation}\label{equ_fano}
\begin{split}
      \inf_{(\hB,\hW)} \sup_{(\Bs, \Ws )}\sum_{t=1}^T\Vert X_t(\hat{B}\hat{w}_t - B^*w_t^*)\Vert_2^2
      &\geq (0.9 \underline{\lambda})^2\inf_{(\hB,\hW)} \sup_{(\Bs, \Ws )}\sum_{t=1}^T\Vert\hat{B}\hat{W} - B^*W^*\Vert_F^2\\
      &\geq \frac{(\delta_1^\prime)^2}{4} \{ 1- \frac{\frac{1}{M^2}\sum_{i,j=1}^M D(\mathbb{P}^{\,i}\,\Vert\  \mathbb{P}^{\,j}) + \ln 2}{\ln M }\}\\
      &=: \frac{(\delta_1^\prime)^2}{4} \cdot C_{Fano}
\end{split}
\end{equation}

Besides, let $c_2\geq1$ be the universal constant in Theorem~\ref{thm_G1}. Note $d, T > k \geq 1$ and thus $\frac{c_2(k(d-k)+kT)}{3} > \frac{2}{3} > \ln2$, we let $(\delta_1^\prime)^2 = \frac{c_2\sigma^2(k(d-k)+kT)}{108\overline{\lambda}^2}$, which enable $C_{Fano} \geq \frac{1}{2}$. Then finally we have:
\begin{equation}
\begin{split}
      \inf_{(\hB,\hW)} \sup_{(\Bs, \Ws )}\sum_{t=1}^T\Vert X_t(\hat{B}\hat{w}_t - B^*w_t^*)\Vert_2^2
      \gtrsim 
      \frac{\sigma^2(k(d-k)+kT)}{\kappa^2} 
\end{split}
\end{equation}
Then from Assumption~\ref{assump_theta1_var} and our notation above, we have $\kappa^2 = \overline{\lambda}/\underline{\lambda} = \Theta(1)$, so we finish the proof.


\qed

\subsection{Main Proof for the ER bound of P/A-MTRL}
\begin{lemma}\label{lemma_nu_bound}
Denote that for any $p \in \mathbb{N}^+$:
\begin{equation}\label{equ_sol_v1}
    \begin{split}
        \nu^p(w_{T+1}^*) &= \arg \min_{\nu}\Vert \nu\Vert_p \qquad s.t. \  W^*\nu = w^*_{T+1}\\
    \end{split}
\end{equation}
and let 
\begin{equation}\label{eq:H_cw}
    H(c_w) = \{w \in \R^{k} | \|w\|_2 = c_w\}
\end{equation}
with constant $c_w > 0$, then for any fixed $W^*$, we have  
\begin{equation}\label{equ_nu_bound}
    \begin{split}
    \sup_{w_{T+1}^* \in H(c_w)}\Vert \nu^p(w_{T+1}^*) \Vert_2 
    & = \frac{c_{w}}{\sigma_{min}(W^*)}
    \\
    \sup_{w_{T+1}^* \in H(c_w)}\Vert \nu^1(w_{T+1}^*) \Vert_1
    &\leq \sqrt{k}\frac{c_{w}}{\sigma_{min}(W^*)}
    \end{split}
\end{equation}
\end{lemma}

\noindent\textbf{Proof of Lemma~\ref{lemma_nu_bound}}.
\paragraph{First equality of ~(\ref{equ_nu_bound})}
Firstly, by definition, we directly have for any $w_{T+1}^*$,
\begin{equation}\label{equ_cw}
    \sigma_{\min}(W^*)\Vert \nu^p(w_{T+1}^*)\Vert_2 \leq \Vert W^* \nu^p(w_{T+1}^*)\Vert_2 = \Vert w_{T+1}^*\Vert_2
\end{equation}
Next we are going to prove the lower bound to show the equality. Let $W^* = UDV^\top$, where $U\in O_{k\times k}, V \in O_{T \times k}, D = diag(\sigma_1(W^*),..., \sigma_k(W^*))$ with $\sigma_1(W^*) > ... > \sigma_k(W^*)$. There always exists an $w'$ satisfies 
\begin{equation}
    \frac{w'}{\Vert w'\Vert_2} = Ue_k
\end{equation}
Then it is easy to see that the corresponding $\nu^p(w')$ satisfies  $V^\top \nu^p(w') =  \Vert w_{T+1}^*\Vert_2\cdot(\sigma_{\min}(W^*))^{-1} e_k$. 
After rearranging, we have
\begin{equation}\label{equ_v2sup}
    \begin{split}
    \frac{\Vert w_{T+1}^*\Vert_2}{\sigma_{\min}(W^*)} 
    = \left\Vert \frac{\Vert w_{T+1}^*\Vert_2}{\sigma_{\min}(W^*)} e_k \right\Vert_2 
    = \Vert V^\top \nu^p(w') \Vert_2 
    \leq \Vert \nu^p(w') \Vert_2 
    \leq \sup_{w_{T+1}^*}\Vert \nu^p(w_{T+1}^*) \Vert_2 
    \end{split}
\end{equation}
Combine (\ref{equ_cw}) and (\ref{equ_v2sup}) we finish the first part.

\paragraph{Second equality of ~(\ref{equ_nu_bound}).} 
It is easy to upper bound
\begin{equation}
\begin{split}
    \Vert\nu^1(w_{T+1}^*)\Vert_1 
    \leq  \sqrt{\Vert\nu^1 (w_{T+1}^*)\Vert_0} \Vert\nu^1 (w_{T+1}^*)\Vert_2
    \leq \sqrt{\Vert \nu^1 (w_{T+1}^*)\Vert_0} \frac{\Vert w_{T+1}^*\Vert_2}{\sigma_{\min}(W)}
\end{split}
\end{equation}
where the last inequality again comes from (\ref{equ_cw}) and the definition $W^* \nu^1(w_{T+1}^*) = w_{T+1}^*$.
Now we can upper bound $ \Vert \nu^1 (w_{T+1}^*)\Vert_0$ by k from the following arguments.
\\\\
Note that the original $l_1$ minimization for the undetermined linear equation $W^*\nu = w_{T+1}^*$
is equivalent to finding the solution to the following linear programming problem.
\begin{equation}
\begin{aligned}\label{equ_LP}
    \min_{\nu_{\pm}} & \ \mathbf{1}^T \nu_{\pm}\\
    \text{s.t.} & \ 
    W_{\pm}\nu_{\pm} = w_{T+1}^*,\\
    & \ \nu_{\pm} \geq 0.
\end{aligned}
\end{equation}
where $\mathbf{1}^\top := (1,...,1) \in \R^{2T}$,
$\nu_{\pm}^\top := (\nu^{+}, \nu^{-})$, $\nu^{+} := \max(\nu, 0)$, $\nu^{-} := \max( -\nu, 0)$ and $W_{\pm} := (W^*, -W^*) \in \R^{k\times 2T}$. Since $W^*\nu^* = w_{T+1}^*$ holds and there exists at least one optimal solution which is a basic feasible solution for LP (\ref{equ_LP}). From Def.~2.9 and Theorem 2.3 in ~\cite{Bertsimas1997IntroductionTL}, we know that the cardinality for the basis of basic feasible solutions is $rank(W_{\pm}) = k$. 
so $\nu^1$ at most $k$-sparse, i.e., $\Vert\nu^1\Vert_0 \leq k$. 

\qed

We show the Lemma that reflects our motivation to get the lower bound of $\|\widetilde \nu^2\|_1^2$.
\begin{lemma}\label{lemma_nu_worst}
Assume conditions in Theorem~\ref{thm_worst_target_ER} hold,
$N_{tot} \rightarrow \infty$, 
and $W^*$ can be any matrix in $\Gamma(\sigma_k) = \{W \in \R^{k\times T} | \sigma_{\min}(W) \geq \sigma_k\}$, then for L2-A-MTRL and P-MTRL we have
\begin{equation}
    \sup_{w_{T+1}^* \in H(c_w)}\Vert \widetilde{\nu}^2(w_{T+1}^*) \Vert_1^2 \gtrsim \frac{T\cdot c_w^2}{N_{tot}\cdot\sigma^2_{\min}(W^*)}
\end{equation}
\end{lemma}

\noindent\textbf{Proof of Lemma~\ref{lemma_nu_worst}.}
For passive learning, actually we can choose any $\nu^p$ such that $W^*\nu^p(w_{T+1}^*) = w_{T+1}^*$, then from Lemma~\ref{lemma_nu_bound} we have:
\begin{equation}
    \sup_{w_{T+1}^* \in H(c_w)}\Vert \widetilde{\nu}^p(w_{T+1}^*) \Vert_1^2
    =  \frac{T}{N_{tot}}\cdot\sup_{w_{T+1}^* \in H(c_w)}\Vert \nu^p(w_{T+1}^*) \Vert_1^2
    = \frac{T\cdot c_w^2}{N_{tot}\cdot\sigma^2_{\min}(W^*)}
\end{equation}
For $L_2$ strategy we have $n_t = \max\{c'' \nu^2(t)^2, \underline{N}\}$. refer to the SVD decomposition of $W^*$ in Lemma~\ref{lemma_nu_bound} and the worst target vector $w^\prime$ defined in (\ref{equ_v2sup}), we have
\begin{equation}
    \nu^2(w^\prime) = VD^{-1}U^\top w^\prime
    = \|w^\prime\|_2\cdot VD^{-1}U^\top Ue_k
    = \|w^\prime\|_2\sigma_{\min}^{-1}(W^*)\cdot V_{*,k}
\end{equation}
where $V_{*,k}$ is the $k$-th column vector of $V \in O_{T,k}$. Since $N_{tot}\gg T\underline{N}$ and $\|\nu^2\|_2 = \|w^\prime\|_2\sigma^{-1}_{\min}(W^*)\|V_{*,k}\|_2^2 = \|w^\prime\|_2\sigma_{\min}^{-1}(W^*)$, then for any $t \in S$, we have
\begin{equation}
    n_t \approx N_{tot} \frac{ |\nu^2(t)|^2}{\|\nu^2\|_2^2} = N_{tot}\cdot V_{t,k}^2
\end{equation}
So as $N_{tot}\rightarrow +\infty$, $t \in S \Leftrightarrow |V_{t,k}| > 0$. 
Note that the minimax lower bound used in Theorem~\ref{thm_worst_target_ER} is proved by using Fano's inequality to the $\delta_V$-separated subset as in Lemma~\ref{lemma_pack}, and the corresponding separated set $G_W$ for $W\in\R^{k\times T}$ is constructed from $G_V$. Clearly $G_{W^\prime}:= \{W \in G_W | W = UDV^\top, \exists t \in [T], \text{s.t.} V_{t,k} = 0\}$ occupy zero volume space in $G_W$, and thus we can use $G_{W}-G_{W^\prime}$ to replace the original $G_W$ set by excluding a corrsponding zero volume space in (\ref{equ_v_cond}) from Lemma~\ref{lemma_pack} which has no difference to the original results. So set $\|w^\prime\|_2 = c_w$, with probability $1-o(1)$ we have $V_{t,k} > 0$ and thus
\begin{equation}
     \sup_{w_{T+1}^* \in H(c_w)}\Vert \widetilde{\nu}^2(w_{T+1}^*) \Vert_1^2 
     \overset{w_{T+1}^* = w^\prime}{\geq} \sum_{t\in S}\frac{|\nu^2(t)|^2}{c''|\nu^2(t)|^2} +  \sum_{t \notin S} \frac{|\nu^2(t)|^2}{\underline{N}}
     \gtrsim \frac{|S|}{c''}
     = \frac{T}{c''}
\end{equation}
where $c'' = N_{tot}\sigma_{\min}^2(W^*)c_{w}^{-2} $.

\qed

We then prove a simple lemma to show that with a particular condition, we have $\|Av\| \approx \|A\|_F \|v\|$.
\begin{lemma}\label{lemma_F_norm}
    Assume $v \in \R^{b}$, $A, \Delta A \in \R^{a\times b}$ and $\|\Delta A\|_F = c \cdot \|A\|$ for some $a,b \in \mathbb{N}^+ $ and $c \in (0, 1)$. Further assume that $A$ satisfies $\|Av\| = \|A\|_F \|v\|$, then
    \begin{equation}
        \|(A+\Delta A)v\| \geq \frac{1-c}{1+c} \|A+\Delta A\|_F \cdot \|v\|
    \end{equation}
\end{lemma}

\noindent\textbf{Proof of Lemma~\ref{lemma_F_norm}.}
We proof it directly:
\begin{equation}
\begin{aligned}
    \|(A+\Delta A)v\| 
    &\geq \|Av\| - \|\Delta A\cdot v \|
    = \|A\|_F\|v\| - \|\Delta A\cdot v \|
    \geq (\|A\|_F - \|\Delta A\|_F)\| v \|\\
    &= \frac{1-c}{1+c}(\|A\|_F+\|\Delta A\|_F) \|v\|
    \geq \frac{1-c}{1+c}\|A+\Delta A\|_F\|v\|
\end{aligned}
\end{equation}
\qed

With such Lemma, we can prove an important Lemma for the lower bound of L2-A-MTRL and P-MTRL.
\begin{lemma}\label{lemma:BWnu_LB_L2}
    Recall the definition of $H(c_w)$ and $\nu^p(w_{T+1}^*)$ in (\ref{equ_sol_v1}) and (\ref{eq:H_cw}), we have the following results for $L_2$-minimization solution.
    \begin{equation}
        \inf_{(\hB, \widetilde{W})} \sup_{(\Bs, \widetilde{W}^*, w_{T+1}^*\in H(c_w))} \|(\hB\widetilde{W} - {B}^*\widetilde{W}^*)\tilde{\nu}^2(w_{T+1}^*)\|_2^2 \gtrsim \sigma^2\cdot k(d-k) \cdot
        \frac{T\cdot c_w^2}{k\cdot N_{tot}\cdot \sigma^2_{\min}(W^*)}
    \end{equation}
\end{lemma}
\noindent\textbf{Proof of Lemma~\ref{lemma:BWnu_LB_L2}.}
The key idea is that we want to find some $\widetilde{W}^*$ such that $\|(\hat{B}-B^*)\widetilde{W}^*\nu^2\| \gtrsim \|(\hat{B}-B^*)\widetilde{W}\|_F  \|\nu^2\|$ when all the row vectors of $\widetilde{W}$ are almost aligned with $\nu^2$. Without loss of generality, we assume $\nu^2(t) \neq 0,~\forall t \in [T]$, and thus when $N_{tot}\rightarrow \infty$, we have $n_t = c'' \cdot |\nu^2(t)|^2,~\forall t \in [T]$, where $c'' \gg 1$ is some constant satisfies $c'' = N_{tot} / \|\nu\|^2$.
We prove the Lemma step by step.

First, we construct a specific $\widetilde{W^*}$, which  is almost rank-1 and has rows aligned with $\widetilde\nu^2$, to achieve the lower bound.
For any given $\nu(t)$, we define
\begin{equation}\label{eq:def_W}
    \widetilde{W^*} := \frac{1}{\sqrt{c''}}u \cdot \chi^\top + \widetilde{\Delta W^*}
\end{equation}
where $u \in \R^T$ is some vector to be determined later and
\begin{equation}\label{eq:def_chi}
    \chi(t) := \text{sgn}(\nu^2(t)) = \mathbb{I}[\nu^2(t)>0] - \mathbb{I}[\nu^2(t)<0], \quad \chi \in \R^{T}
\end{equation}
\begin{equation}\label{eq:def_Delta_W}
    \widetilde{\Delta W^*}:= \sum_{i=2}^k \widetilde{\sigma}_i \widetilde{\alpha}_i \widetilde{\beta}_i^\top
\end{equation}
Obviously, $\|\chi\| = \sqrt{T}$.
Here $\widetilde\alpha_i, \widetilde\beta_i \in \R^{T},~\forall i \in \{2,\ldots,k\}$ and we let $\{u/\|u\|_2,\widetilde\alpha_2,\ldots,\widetilde\alpha_k\}$ and $\{\chi/\sqrt{T},\widetilde\beta_2,\ldots,\widetilde\beta_k\}$ to be two orthonormal bases of two $k$-dimensional subspace of $\R^T$. The reason for such a definition of $\widetilde{\Delta W^*}$ is that the Eqn.~\ref{eq:def_W} will naturally be an SVD form of $\widetilde{W}^*$. And for simplicity, we let $\widetilde{\sigma}_i \equiv \widetilde{\sigma}_k,~\forall i \in \{2,\ldots,k\}$. We then let
\begin{equation}\label{eq:rank1_dominate}
    \begin{split}
        \|\frac{1}{\sqrt{c''}}u\cdot \chi^\top\|_F = 2\|\widetilde{\Delta W^*}\|_F ~\Leftrightarrow ~ \widetilde\sigma_k = \frac{\|u\|\sqrt{T}}{2\sqrt{(k-1)c''}}
    \end{split}
\end{equation}
Then from $\widetilde \nu^2(t) = \nu^2(t)/\sqrt{n_t} = \chi(t) / \sqrt{c''}$ and $\widetilde{w}^*(t) = \sqrt{n_t}w^*(t) = \sqrt{c''}|\nu^2(t)|\cdot w^*(t)$, 
we have
\begin{equation}
    \begin{split}
        W^*\nu^2 = \widetilde{W^*}\widetilde{\nu}^2=  \frac{T}{c''}u + \frac{1}{\sqrt{c''}}\sum_{i=2}^k \widetilde \sigma_k \widetilde \alpha_i
        \widetilde\beta_i^\top \chi = \frac{T}{c''}u
    \end{split}
\end{equation}
Note that $W^*\nu^2 = w_{T+1}^* \in H(c_w)$, i.e., $c_w = \|W^*\nu^2\|$, we have the following conditions for $\|u\|$ and $\widetilde \sigma_k$.
\begin{equation}
    \begin{split}
        \|u\|_2 = \frac{c_w \cdot c''}{T}, ~\widetilde\sigma_k = \frac{c_w\sqrt{c''}}{2\sqrt{(k-1)T}}
    \end{split}
\end{equation}

We then choose $\nu^2 = \nu' := \mathbf{1} = [1,\ldots,1]^\top \in \R^T$, thus $\chi = \nu' = \mathbf{1}, \|\nu'\| = \sqrt{T}$ and for $W^*$ we have:
\begin{equation}
\begin{split}
    W^*
    = \frac{u}{c''}\mathbf{1}^\top + \frac{1}{\sqrt{c''}}\widetilde\sigma_k\sum_{i=2}^k \widetilde \alpha_i \widetilde\beta_i^\top 
    =\frac{1}{\sqrt{c''}}\cdot \widetilde{W^*}
\end{split}
\end{equation}
And thus
\begin{equation}
    \sigma_{\min}(W^*) = \sigma_k = \frac{\widetilde\sigma_k}{\sqrt{c''}} = \frac{c_w}{2\sqrt{(k-1)T}}
\end{equation}
which results that $T = \|\nu'\|^2 > c_w^2[4(k-1)\sigma_k^2]^{-1}$.
And we get
\begin{equation}\label{eq:nup_LB}
    \|\widetilde{\nu'}\|^2 = \sum_{t=1}^T \frac{|\nu'(t)|^2}{c''|\nu(t)|^2} = \frac{T\|\nu_2\|^2}{N_{tot}} 
    \gtrsim
    \frac{Tc_w^2}{k\sigma_k^2N_{tot}}
\end{equation}
We check that $\nu'$ is a valid choice for the $L_2$-minimization solution.
Let $W^* = UDV^\top$ be the SVD form of $W^*$, where $U = (u/\|u\|, \widetilde \alpha_2,\ldots,\widetilde\alpha_k) \in O_{k\times k}, V = (\chi/\sqrt{T},\widetilde \beta_2, \ldots, \widetilde \beta_k ) \in O_{T\times k}, D = \text{diag}(\sigma_1,\ldots,\sigma_k), \sigma_1 \geq \ldots \geq \sigma_k \geq 0$. 
Note that
\begin{equation}
VD^{-1}U^{\top}W^*\nu' = VV^\top \nu' = (\frac{1}{T}\chi \chi^\top + \sum_{i=2}^k \widetilde \beta_i \widetilde \beta_i^\top)\nu' = \chi + 0 = \nu'. 
\end{equation}
Therefore, $\nu' = \arg\min_{W^*\nu' = W^*x}\|x\|_2$.
Here we use the fact that $\widetilde \beta_i^\top \nu' = \widetilde \beta_i^\top \chi = 0,~\forall i \in \{2,\ldots,k\}$.

Finally we have:
\begin{equation}
    \begin{split}
        &\inf_{(\hB, \widetilde{W})} \sup_{(\Bs, \widetilde{W}^*, w_{T+1}^*\in H(c_w))} \|(\hB\widetilde{W} - {B}^*\widetilde{W}^*)\tilde{\nu}^2(w_{T+1}^*)\|_2^2\\
        &\gtrsim
        \inf_{\hB} \sup_{\Bs} \|(\hB - {B}^*)\widetilde{W}^*\tilde{\nu'}\|_2^2\\
        &\gtrsim  \inf_{\hB} \sup_{\Bs} \|(\hB - {B}^*)\widetilde{W}^*\|_F^2\|\tilde{\nu}'\|_2^2,\quad (\text{Eqn}.~\ref{eq:rank1_dominate}, \text{Lemma~\ref{lemma_F_norm}})\\
        &\asymp \inf_{\hB} \sup_{\Bs} \sum_{t=1}^T \|X_t(\hB - {B}^*){w}_t^*\|_2^2 \cdot \frac{T c_w^2}{k \sigma_k^2 N_{tot}},\quad (\text{Eqn.~\ref{eq:nup_LB}})\\
        &\gtrsim \sigma^2 \cdot k(d-k)\cdot \frac{T c_w^2}{k \sigma_k^2 N_{tot}}
    \end{split}
\end{equation}
For the first and last inequality, we restrict the local packing space on $W$ and obtain the results in a manner similar to Theorem~\ref{thm_XBW_LB}. Specifically, we note that the orthonormal matrix $B$ can be viewed as a Grassmann manifold that is diffeomorphic to a $k\times(d-k)$ dimensional linear matrix\cite{bai1992preliminary}, and the constraint $Bw = \mathbf{0}$ introduces at most $d$ additional limiting equations to $B$, which will not influence its local packing number. Therefore, it becomes straightforward to prove the last inequality using a methodology similar to the proof of Theorem~\ref{thm_XBW_LB}. And we finish the proof.

\qed

Finally, we turn to our main theorem in Sec.~\ref{sec: minimax bound}.

\noindent\textbf{Proof of Theorem~\ref{thm_worst_target_ER}.}
From the conditions, we have $c_w = \Theta(1)$.
\paragraph{Upper bound of ER for L1-A-MTRL.}
From Eqn.~\ref{eq:c_and_N_tot} from the proof of Lemma~\ref{lemma_nu}, we get $\|\widetilde{\nu}^1\| \leq (1+o(1)) \|\nu^1\|_1^2 / N_{tot}$ when $N_{tot} \gg T\underline{N}$.
Then use the second inequality of (\ref{equ_nu_bound}) in Lemma~\ref{lemma_nu_bound}, we have
\begin{equation}\label{inequ_worst_v1_sup_nu}
    \sup_{w_{T+1}^* \in H(c_w)}\Vert \widetilde{\nu}^1(w_{T+1}^*) \Vert_1^2 \lesssim \sup_{w_{T+1}^* \in H(c_w)} \frac{\|\nu^1(w_{T+1})\|_1^2}{N_{tot}} \leq \frac{k\cdot c_w^2}{N_{tot}\cdot\sigma^2_{\min}(W^*)}
\end{equation}
For the upper bound, let $\widetilde{w}_{t} = \hat{w}_t\sqrt{n_t}$, $\widetilde{w}^*_{t} = \hat{w}^*_t\sqrt{n_t}$ and $\widetilde{\nu}^2(t) = \frac{\nu^*(t)}{\sqrt{n_t}}$ for all $t \in [T]$, then we have:
\begin{equation}\label{inequ_worst_UB_v1}
    \begin{aligned}
        \E_{x \sim \mu_{T+1}} \Vert x^\top(\hat{B}\hat{w}_{T+1} - B^*w_{T+1}^*) \Vert_2^2
        &= 
        \Vert (\Sigma_{T+1}^*)^{\frac{1}{2}}(\hat{B}\hat{W} - B^*W^*)\nu^1 \Vert_2^2\\
        &\leq \| (\Sigma_{T+1}^*)^{\frac{1}{2}}(\hat{B}\widetilde{W} - B^*\widetilde{W}^*)\|_F^2 \cdot \|\widetilde{\nu}^1 \|^2\\ 
        &= \sum_{t=1}^Tn_t\|(\Sigma_{T+1}^*)^{\frac{1}{2}}(\hat{B}\hat{w}_t - B^*w^*_t)\|^2 \cdot \|\widetilde{\nu}^1 \|^2\\
        &\asymp \sum_{t=1}^T n_t\|(\Sigma_{t}^*)^{\frac{1}{2}}(\hat{B}\hat{w}_t - B^*w^*_t)\|^2 \cdot \|\widetilde{\nu}^1 \|^2, \qquad(\text{Assumption~\ref{assump_theta1_var}})\\
        & \lesssim \sum_{t=1}^T\|X_{t}(\hat{B}\hat{w}_t - B^*w^*_t)\|^2 \cdot \|\widetilde{\nu}^1 \|^2, \qquad (\text{Lemma~\ref{lemma_sigma_kd}})\\
        & \leq \sigma^2(kd\ln(\frac{N_{tot}}{T}) + kT +\ln(\frac{1}{\delta})) \|\widetilde{\nu}^1 \|^2,\quad (\text{Claim C.1 in \cite{chen2022active}})
    \end{aligned}
\end{equation}

Then combine (\ref{inequ_worst_UB_v1}) and (\ref{inequ_worst_v1_sup_nu})  we prove the result for L1-A-MTRL.

\paragraph{Lower bound of ER for P-MTRL/L2-A-MTRL.}
For L2-A-MTRL, we derive the results from Lemma~\ref{lemma:BWnu_LB_L2}. It can be easily verified that the same results hold for P-MTRL since we set $\nu' = [1,\ldots,1]^\top \in \R^T$ in Lemma~\ref{lemma:BWnu_LB_L2}.
\qed


\section{Proof of Theorem~\ref{thm_noise}}\label{appendix_noise_L1}
Before proofing the original Theorem, we first illustrate an assumption naturally used for the sparse linear model and Lasso Program~\cite{wainwright_2019}:
\begin{assumption}\label{assump_RE}
(RE condition) Let $\nu^*$ be supported on a subset $S \in [T]$ with $|S| = s$ (From Theorem~\ref{thm_worst_target_ER} we know $s \leq k$). Then
$W^*$ satisfies \textit{Restricted Eigenvalue} condition over S with parameters ($\kappa, 3$) if: 
\begin{equation}\label{inequ_W_kappa}
    \Vert W^* \Delta\Vert_2^2 \geq \kappa \Vert  \Delta\Vert_2^2, \qquad \forall \Delta \in \mathbb{C}_{3}(S)
\end{equation}
where $\mathbb{C}_{\alpha}(S) :=\{\Delta \in \R^k | \Vert\Delta_{S^c}\Vert_1 \leq \alpha\Vert\Delta_{S}\Vert_1 \}$.
\end{assumption}
What should be mentioned is that in this section we just consider L1-A-MTRL, so we replace $\hat{\nu}$ and $\nu^*$ with $\hat{\nu}^1$ and $\nu^1$, respectively.

Since $\sigma_{\max}^2(W^*)\geq \kappa \geq \sigma_{\min}^2(W^*)$, we rewrite Theorem~\ref{thm_noise} with RE condition as follows. Once we prove the following theorem, we can replace $\kappa$ with $\sigma_{\min}^2(W^*)$ and $\sigma_{\max}^2(W^*)$ correspondingly and immediately prove the original theorem.
\begin{theorem}\label{thm_noise_rewrite}
Let Assumption~
~\ref{assump_subG},
\ref{assump_dimension}, \ref{assump_W_fullr},
\ref{assump_bounded_W},
\ref{assump_Sigma},
\ref{assump_RE}
hold. 
Let $\Lambda$ denote the lower bound of $\|\nu^*\|_1$, $q = \frac{\sqrt{k}R}{\usigma}$ (so $q \geq \|\nu^*\|_1$) and
$\gamma \geq \max\{2160sqC_W\Lambda^{-1}, \sqrt{2160sq\kappa\Lambda^{-1}}\}$
and $\underline{\sigma} = \sigma_{\min}(W^*) >0$. Then in order to let $ER_{L_1}\leq \varepsilon^2$ with probability $1-\delta$, the number of source samples $N_{total}$ is at least
\begin{equation}\label{equ_O_order}
\mathcal{\widetilde{O}}(\sigma^2(kd + kT )\Vert\nu^*\Vert_1^2\varepsilon^{-2} + T\beta)
\end{equation}
where $\beta = \max\{\gamma^2\frac{\sigma_z^2}{\kappa^2},\gamma^2\frac{C_W^2}{\kappa^2}\rho^4,\rho^4, \frac{\sigma_z^2}{\kappa}\}\cdot(d + \ln(\frac{4T}{\delta}))$, and target task sample complexity $n_{T+1}$ is at least
\begin{equation}
\mathcal{\widetilde{O}}(
\sigma^2k\varepsilon^{-2}
+ \alpha)
\end{equation}
where $\alpha = \max\{ \gamma^2\frac{\sigma_z^2}{\kappa^2\Lambda^2}, \gamma^2\frac{C_W^2}{\kappa^2}\rho^4, \rho^4\}
\cdot(k + \ln(\frac{4}{\delta}))$.
\end{theorem}
\begin{lemma}\label{lemma_RE}
(A variant of Theorem 7.13 in \cite{wainwright_2019}) Assume that Assumption
\ref{assump_RE} hold. Any  solution of the Lagrangian Lasso (\ref{equ_Lasso}) with regularization parameter lower bounded as $\lambda_k \geq 2 \Vert \hat{W}^\top z\Vert_{\infty}$ satisfies the following bound
\begin{equation}
    \Vert \hat{\nu} - \nu^*\Vert_2 \leq \frac{3}{\kappa}\sqrt{s}\lambda_k
\end{equation}
\begin{equation}
    \Vert \hat{\nu} - \nu^*\Vert_1 \leq 
    4\sqrt{s}\Vert \hat{\nu} - \nu^*\Vert_2
\end{equation}
\end{lemma}
\qed

\begin{remark}
In Theorem~\ref{thm_nu_hat_bound} we want $\epsilon \leq \min(0.05, \frac{\kappa}{4\gamma C_W})$ with high probability, so from Lemma~\ref{lemma_sigma_kd}, we need 
$n_t > \max(400, \frac{16\gamma^2C_W^2}{\kappa^2})c_a\rho^4(d + \ln(\frac{2T}{\delta}))$ for all $t\in[T]$ and $n_{T+1} > \max(400, \frac{16\gamma^2C_W^2}{\kappa^2})c_a\rho^4(k + \ln(\frac{2}{\delta}))$ for universal constant $c_a > 0$.
\end{remark}
To get the bound of regularization parameter $\lambda_k$, we turn to control the bound of the noise term $z$ since $\hat{W}$ and $\hat{w}_{T+1}^*$ are solved by original least square method.
\begin{theorem}\label{thm_nu_hat_bound}
If $n^i_t \geq \max\{3\gamma^2\frac{\sigma_z^2}{\kappa^2},16\gamma^2\frac{C_W^2}{\kappa^2}c_a\rho^4,400c_a\rho^4, \frac{12\sigma_z^2}{\kappa}\}\cdot(d + \ln(\frac{4T}{\delta}))$, $n^i_{M+1} \geq 
\max\{ 3\gamma^2\frac{\sigma_z^2}{\kappa^2\Vert\nu^*\Vert_1^2}, \\16\gamma^2\frac{C_W^2}{\kappa^2}c_a\rho^4, 400c_a\rho^4\}
\cdot(k + \ln(\frac{4}{\delta}))$,
and 
Assumption
~\ref{assump_RE}
,~\ref{assump_bounded_W}
,~\ref{assump_Sigma} 
hold. Then with probability $1-\delta$ we have
\begin{equation}\label{inequ_v_v*}
    \Vert \hat{\nu} - \nu^*\Vert_1 \leq \frac{2160}{\gamma}s  \cdot\max\{C_W,\frac{\kappa}{\gamma}\}
    \cdot \frac{\sqrt{k}R}{\usigma}
\end{equation}
\end{theorem}
\begin{remark}
If (\ref{inequ_v_v*}) holds and $\frac{\sqrt{k}R}{\usigma} = \Theta( \|\nu^*\|_1)$, 
then active learning method with L1-minimization just multiplies an additional term $1+\frac{2160}{\gamma}s  \max\{C_W,\frac{\kappa}{\gamma}\}$, i.e.
\begin{equation}\label{inequ_practical_L1_UB}
    ER_{active}  \lesssim \sigma^2(kd\ln(\frac{N_{tot}}{T}) + kT )\frac{\Vert\nu^*\Vert_1^2}{N_{tot}}(1+\frac{2160}{\gamma}s  \max\{C_W,\frac{\kappa}{\gamma}\})^2 + \sigma^2\frac{(k +  \ln(\frac{1}{\delta}))}{n_{T+1}}
\end{equation}
\end{remark}
\noindent\textbf{Proof of  Theorem~\ref{thm_nu_hat_bound}.} 

\textbf{Substep 1: Decompose $z$.}

As the analysis of original least square method in ~\cite{chen2022active},  for every $t\in[T+1]$ we have:
\begin{equation}\label{equ_wti}
\begin{split}
    \hat{w}_t^i &= \arg\min_{w} \Vert X_t^i\hat{B}^iw - Y_t\Vert_2\\
    &=((X_t^i\hat{B}^i)^\top X_t^i\hat{B}^i)^{-1}(X_t^i\hat{B}^i)^\top Y_t\\
    &=((\hat{B}^i)^\top\hat{\Sigma}_{t}^i \hat{B}^i)^{-1}(\hat{B}^i)^\top\hat{\Sigma}_{t}^i B^*w_{t}^* + \frac{1}{n_t}((\hat{B}^i)^\top\hat{\Sigma}_{t}^i \hat{B}^i)^{-1}(\hat{B}^i)^\top (X_t^i)^\top Z_t \\
\end{split}
\end{equation}

Then we have
\begin{equation}\label{equ_zi_decompose}
\begin{split}
    z^i =& \hat{w}_{T+1}^i - \hat{W}^i\nu^*\\
    =& \hat{w}_{T+1}^i - \sum_{t=1}^T\hat{w}_t^i\nu_t^*\\
    =& ((\hat{B}^i)^\top\hat{\Sigma}_{T+1}^i \hat{B}^i)^{-1}(\hat{B}^i)^\top\hat{\Sigma}_{T+1}^i B^*w_{T+1}^* - \sum_{t=1}^T ((\hat{B}^i)^\top\hat{\Sigma}_{t}^i \hat{B}^i)^{-1}(\hat{B}^i)^\top\hat{\Sigma}_{t}^i B^*w_{t}^*\nu^*_{t}\\
    &+ \frac{1}{n_{T+1}}((\hat{B}^i)^\top\hat{\Sigma}_{T+1}^i \hat{B}^i)^{-1}(\hat{B}^i)^\top (X_{T+1}^i)^\top Z_{T+1} - \sum_{t=1}^T\frac{1}{n_t}((\hat{B}^i)^\top\hat{\Sigma}_{t}^i \hat{B}^i)^{-1}(\hat{B}^i)^\top (X_t^i)^\top Z_t\nu^*_t\\
    \\
    =& \underbrace{((\hat{B}^i)^\top\hat{\Sigma}_{T+1}^i \hat{B}^i)^{-1}(\hat{B}^i)^\top\hat{\Sigma}_{T+1}^i B^*w_{T+1}^* - ((\hat{B}^i)^\top\Sigma^* \hat{B}^i)^{-1}(\hat{B}^i)^\top\Sigma^* B^*w_{T+1}^* }_{E_1^i}\\ 
    &-\underbrace{(\sum_{t=1}^T ((\hat{B}^i)^\top\hat{\Sigma}_{t}^i \hat{B}^i)^{-1}(\hat{B}^i)^\top\hat{\Sigma}_{t}^i B^*w_{t}^*\nu^*_{t} - \sum_{t=1}^T ((\hat{B}^i)^\top\Sigma^* \hat{B}^i)^{-1}(\hat{B}^i)^\top\Sigma^* B^*w_{t}^*\nu^*_{t})}_{E_2^i}\\
    &+ \underbrace{\frac{1}{n_{T+1}}((\hat{B}^i)^\top\hat{\Sigma}_{T+1}^i \hat{B}^i)^{-1}(\hat{B}^i)^\top (X_{T+1}^i)^\top Z_{T+1} }_{E_3^i}
    - \underbrace{\sum_{t=1}^T\frac{1}{n_t}((\hat{B}^i)^\top\hat{\Sigma}_{t}^i \hat{B}^i)^{-1}(\hat{B}^i)^\top (X_t^i)^\top Z_t\nu^*_t}_{E_4^i}\\
\end{split}
\end{equation}

where the third equality of Equ.~\ref{equ_zi_decompose} use Equ.~\ref{equ_wti} and the fourth equality  comes from $w_{T+1}^* = W^*\nu^*$. It's obvious that $E_{k}^i, k \in \{1,2,3,4\}$ all have 0 expectation, and to control the bound of $z$, we just need to bound these 4 term in $l_2$-norm for all $i$ and use the inequality
\begin{equation}\label{inequ_z_2}
    \Vert z\Vert_2 = \Vert E_1^i - E_2^i + E_3^i - E_4^i\Vert_2 
    \leq 2 (\Vert E_1^i \Vert_2+ \Vert E_2^i\Vert_2 + \Vert E_3^i\Vert_2 + \Vert E_4^i\Vert_2)
\end{equation}

\textbf{Substep 2: Calculate Error Terms $E_*^i$.}

For the first term, with Inequ.~\ref{inequ_BSigmaB} and Assumption~\ref{assump_Sigma} we have
\begin{equation}\label{inequ_E1}
    \begin{split}
        \Vert E_1^i \Vert_2 &\leq \Vert((\hat{B}^i)^\top\hat{\Sigma}_{T+1}^i \hat{B}^i)^{-1}(\hat{B}^i)^\top\hat{\Sigma}_{T+1}^i B^* - ((\hat{B}^i)^\top\Sigma^* \hat{B}^i)^{-1}(\hat{B}^i)^\top\Sigma^* B^*\Vert_2 \Vert w_{T+1}^* \Vert_2\\
        &\leq \Vert w^*_{T+1}\Vert_2\cdot
        \Vert \frac{1+2\epsilon}{1-2\epsilon}((\hat{B}^i)^\top\Sigma^* \hat{B}^i)^{-1}(\hat{B}^i)^\top\Sigma^* B^* - ((\hat{B}^i)^\top\Sigma^* \hat{B}^i)^{-1}(\hat{B}^i)^\top\Sigma^* B^*\Vert_2 \\
        &\leq 
        \Vert w^*_{T+1}\Vert_2 \frac{4\epsilon}{1-2\epsilon} \Vert ((\hat{B}^i)^\top \hat{B}^i)^{-1}(\hat{B}^i)^\top B^*\Vert_2\\ 
        &\leq  \frac{4\epsilon }{1-2\epsilon}\Vert w^*_{T+1}\Vert_2, \qquad(\sigma_{\max}((\hat{B}^i)^\top B^*) \leq 1)\\
        &\leq \frac{4\epsilon }{1-2\epsilon}C_W\Vert \nu^*\Vert_1, \qquad(\Vert w_{T+1}^*\Vert_2=\Vert\sum_{t=1}^TW^*e_t\nu^*_t\Vert_2 \leq \max_t\Vert W^*e_t\Vert_2\cdot\Vert \nu^*\Vert_1)
    \end{split}
\end{equation}

The fourth inequality is relevant to subspace angle distance between $p$ and $q$, where $\hat{B}^i$ and $B^*$ are orthonormal matrices whose colums form orthonormal bases of $p$ and $q$  respectively, as section 2 in ~\cite{Tripuraneni2021ProvableMO}. The second term $E_2^i$ has upper bound similar to $E_1^i$:
\begin{equation}\label{inequ_E2}
    \begin{split}
        \Vert E_2^i \Vert_2 &\leq \sum_{t=1}^T \Vert ((\hat{B}^i)^\top\hat{\Sigma}_{t}^i \hat{B}^i)^{-1}(\hat{B}^i)^\top\hat{\Sigma}_{t}^i B^* - ((\hat{B}^i)^\top\Sigma^* \hat{B}^i)^{-1}(\hat{B}^i)^\top\Sigma^* B^*\Vert_2 \Vert w_t^*\nu_t^*\Vert_2\\
        &\leq \frac{4\epsilon }{1-2\epsilon}\Vert ((\hat{B}^i)^\top \hat{B}^i)^{-1}(\hat{B}^i)^\top B^*\Vert_2 \sum_{t=1}^T \Vert w_t^*\nu_t^*\Vert_2\\
        &\leq  
        \frac{4\epsilon }{1-2\epsilon}C_W\Vert \nu^*\Vert_1
    \end{split}
\end{equation}
For the third term, from Lemma~\ref{lemma_BXZ} with probability at least $1-\frac{\delta}{4}$ we have:
\begin{equation}\label{inequ_E3}
    \begin{split}
        \Vert E_3^i\Vert_2 
        &\leq \frac{1}{n_{T+1}}\Vert ((\hat{B}^i)^\top\hat{\Sigma}_{T+1}^i \hat{B}^i)^{-1}(\hat{B}^i)^\top (X_{T+1}^i)^\top Z_{T+1}\Vert_2\\
        &\leq \frac{1}{n_{T+1}\cdot (1-2\epsilon)}\Vert ((\hat{B}^i)^\top\Sigma^* \hat{B}^i)^{-1} \Vert_2 \Vert (\hat{B}^i)^\top(X_{T+1}^i)^\top Z_{T+1}\Vert_2\\
        &\leq \frac{\sqrt{1+2\epsilon}}{1-2\epsilon}\cdot \sigma_z \sqrt{\frac{2k+3\ln(\frac{4}{\delta})}{n_{T+1}}}\\
    \end{split}
\end{equation}
Analogously, from Lemma~\ref{lemma_BXZ} with probability at least $1-\frac{\delta}{4}$ we have:
\begin{equation}\label{inequ_E4}
    \begin{split}
        \Vert E_4^i\Vert_2 
        &\leq \sum_{t=1}^T\frac{1}{n_{t}}\Vert ((\hat{B}^i)^\top\hat{\Sigma}_{t}^i \hat{B}^i)^{-1}(\hat{B}^i)^\top (X_{t}^i)^\top Z_{t}\nu_t^*\Vert_2\\
        &\leq \sum_{t=1}^T\frac{1}{n_{t}}\Vert ((\hat{B}^i)^\top\hat{\Sigma}_{t}^i \hat{B}^i)^{-1}(\hat{B}^i)^\top \Vert_2 \Vert (X_{t}^i)^\top Z_{t}\Vert_2 |\nu_t^*|\\
        &\leq \sum_{t=1}^T \frac{\sqrt{1+2\epsilon}}{1-2\epsilon}\cdot\sigma_z\sqrt{\frac{2d+3\ln(\frac{4T}{\delta})}{n_t}}|\nu_t^*|\\
        &\leq \frac{\sqrt{1+2\epsilon}}{1-2\epsilon}\cdot\sigma_z\sqrt{\frac{2d+3\ln(\frac{4T}{\delta})}{\min_t(n_t)}}\Vert \nu^*\Vert_1\\
    \end{split}
\end{equation}

\textbf{Substep 3: Final Calculation.}

Combining (\ref{inequ_E1}), (\ref{inequ_E2}), (\ref{inequ_E3}), (\ref{inequ_E4}) and (\ref{inequ_z_2}), with probability at least $1-\delta$ we have
\begin{equation}\label{inequ_z_E1234}\begin{split}
    \Vert z^i\Vert_2 &\leq 
    \frac{16\epsilon }{1-2\epsilon}C_W\Vert \nu^*\Vert_1
    + \frac{2\sqrt{1+2\epsilon}}{1-2\epsilon}\cdot \sigma_z (\sqrt{\frac{2k+3\ln(\frac{4}{\delta})}{n_{T+1}}} + \sqrt{\frac{2d+3\ln(\frac{4T}{\delta})}{\min_t(n_t)}}\Vert \nu^*\Vert_1)\\
    &\leq \frac{16}{0.9\times 4 \times \gamma}\kappa \Vert\nu^*\Vert_1 + \frac{2\sqrt{1.1}}{0.9}\times\frac{\kappa\Vert\nu^*\Vert_1}{\gamma}\times 2 ,\qquad(
    \text{Conditions})\\
    &\leq \frac{82}{9}\frac{\kappa\Vert\nu^*\Vert_1}{\gamma}
\end{split}
\end{equation}
Choose that
\begin{equation}\label{equ_lambda}
\begin{split}
    \lambda_k 
    &:= 45 \frac{\kappa\sqrt{k}R}{\gamma\underline{\sigma}}\max\{C_W,\frac{\kappa}{\gamma}\} \\
    &\geq 45 \frac{\kappa\Vert\nu^*\Vert_1}{\gamma}\max\{C_W,\frac{\kappa}{\gamma}\} \\
    &\geq 2\times\frac{22}{9} \max\{C_W,\frac{\kappa}{\gamma}\} \times \frac{82}{9}\frac{\kappa\Vert\nu^*\Vert_1}{\gamma} \\
    &\geq 2 \cdot (\max_{t}\Vert \hat{w}_t^i\Vert_2)\cdot\Vert z^i\Vert_2 , \qquad((\ref{inequ_z_E1234}), (\ref{inequ_hat_w}))\\
    &
    \geq  
    2 \max_{t}| (\hat{w}_t^i)^\top z^i| \geq 2 \Vert \hat{W}^\top z^i \Vert_{\infty}
\end{split}
\end{equation}

Finally from Lemma~\ref{lemma_RE}
, the solution of (\ref{equ_Lasso}) with regularization parameter $\lambda_k$ satisfies:
\begin{equation}
    \begin{split}
       \Vert\hat{\nu} - \nu^*\Vert_1 &\leq \frac{12s}{\frac{1}{4}\kappa}\lambda_k, \qquad(\text{Lemma~\ref{lemma_RE},~\ref{lemma_W_hat_RE}})\\
      &=\frac{2160}{\gamma}s\cdot \frac{\sqrt{k}R}{\usigma}\cdot  \max\{C_W,\frac{\kappa}{\gamma}\}, \qquad(\ref{equ_lambda})\\
    \end{split}
\end{equation}
\qed

\begin{lemma}\label{lemma_hat_W}
Assume conditions in Theorem~\ref{thm_nu_hat_bound} hold, then the norms of column vectors of $\hat{W}$ have similar uppper bound to that of $W^*$:
\begin{equation}\label{inequ_hat_w}
    \Vert \hat{w}_t^i\Vert_2 \leq \frac{22}{9}\max\{C_W,\frac{\kappa}{\gamma}\}
\end{equation}
\end{lemma}
\noindent\textbf{Proof of  Lemma~\ref{lemma_hat_W}.} 
This can be done by directly calculation as (\ref{inequ_E2}) and (\ref{inequ_E4})
\begin{equation}
    \begin{split}
      \Vert\hat{w}_t^i\Vert_2 
      &= \Vert((\hat{B}^i)^\top\hat{\Sigma}_{t}^i \hat{B}^i)^{-1}(\hat{B}^i)^\top\hat{\Sigma}_{t}^i B^*w_{t}^* + \frac{1}{n_t}((\hat{B}^i)^\top\hat{\Sigma}_{t}^i \hat{B}^i)^{-1}(\hat{B}^i)^\top (X_t^i)^\top Z_t\Vert_2\\
      &\leq \Vert((\hat{B}^i)^\top\hat{\Sigma}_{t}^i \hat{B}^i)^{-1}(\hat{B}^i)^\top\hat{\Sigma}_{t}^i B^*\Vert_2 \Vert w_{t}^*\Vert_2 + \frac{1}{n_t}\Vert((\hat{B}^i)^\top\hat{\Sigma}_{t}^i \hat{B}^i)^{-1}(\hat{B}^i)^\top \Vert_2 \Vert(X_t^i)^\top Z_t\Vert_2\\
      &\leq \frac{1+2\epsilon}{1-2\epsilon}C_W + \frac{\sqrt{1+2\epsilon}}{1-2\epsilon}\cdot\frac{\kappa}{\gamma}\\
      &\leq \frac{1.1\times 2}{9}\max\{C_W,\frac{\kappa}{\gamma}\}
    \end{split}
\end{equation}
\qed
\begin{lemma}\label{lemma_BXZ}
Assume Assuption~\ref{assump_Sigma} holds. For any $t \in [T]$, with probability $1-\frac{\delta}{4}$ we have
\begin{equation}\label{inequ_XZ}
    \Vert (X^i_t)^\top Z_t\Vert_2 \leq \sigma_z \sqrt{n_{t}(1+2\epsilon)(2d+3\ln(\frac{4T}{\delta}))}
\end{equation}
As for target task, for any  $B \in \R^{d\times k}$ that is independent of $Z_{T+1}$
, with probability $1-\frac{\delta}{4}$ we have
\begin{equation}\label{inequ_BXZ}
    \Vert B^\top (X^i_{T+1})^\top Z_{T+1}\Vert_2 \leq \sigma_z \sqrt{n_{T+1}(1+2\epsilon)(2k+3\ln(\frac{4}{\delta}))}
\end{equation}
\end{lemma}

\noindent\textbf{Proof of  Lemma~\ref{lemma_BXZ}.} We firstly proof ~\ref{inequ_BXZ}. Using SVD we have $B^\top(X_{T+1}^i)^\top = U_{BX}D_{BX}V^\top_{BX}$, where $U_{BX} \in O_{k\times k}, V_{BX} \in O_{n\times k}, D_{VX}= diag(\sigma_1( B^\top (X^i_{T+1})^\top),...,\sigma_k( B^\top (X^i_{T+1})^\top))$. Let $Q := V_{BX}^\top Z_{T+1}$, we know $Q ~\sim \mathcal{N}(0, \sigma_z^2 I_{k})$ since $B, X_{T+1}^i$ are independent to $Z_{T+1}$, so does $V_{BX}$. Note that $\frac{1}{\sigma_z^2}\Vert Q\Vert_2^2 \sim \chi^2(k)$, and thus with probability at least $1-\frac{\delta}{4}$ we have ~\cite{Massart2000}
\begin{equation}\label{inequ_Q}
    \frac{1}{\sigma_z^2}\Vert Q\Vert_2^2 \leq k + 2\sqrt{k\ln{\frac{4}{\delta}}}+2\ln{\frac{4}{\delta}}
\end{equation}
Then use (\ref{inequ_Q}), with probability at least $1-\frac{\delta}{4}$ we have
\begin{equation}\begin{split}
     \Vert B^\top (X^i_{T+1})^\top Z_{T+1}\Vert_2^2
     &= Z_{T+1}^\top  (X^i_{T+1})  B B^\top (X^i_{T+1})^\top Z_{T+1}\\
     &= Z_{T+1}^\top V_{BX}D_{BX}^2V_{BX}^\top Z_{T+1}\\
     &=\sum_{j=1}^k \sigma^2_{j}(B^\top (X^i_{T+1})^\top) Q_j^2\\
     &\leq \sigma_{\max}((X^i_{T+1})^\top X^i_{T+1}) \Vert Q \Vert_2^2\\
     &\leq n_{T+1} \cdot (1+2\epsilon) \cdot \sigma_z^2(2k + 3\ln(\frac{4}{\delta})), \qquad (\text{Assumption}~\ref{assump_Sigma}~, (\ref{inequ_Q}))
    \end{split}
\end{equation}
As for source tasks, (\ref{inequ_BXZ}) don't hold since $\hat{B^i}$ is not independent to $X_{t}^i$ and $Z_{t}$. Then in order to get (\ref{inequ_XZ}), we just need to note that $rank(X_t^i) = d$ and others steps are similar to the proof above.

\qed

\begin{lemma}\label{lemma_W_hat_RE}
If all the conditions of Theorem~\ref{thm_nu_hat_bound} hold, then $\hat{W}$ satisfies RE conditions with parameter $(\frac{1}{4}\kappa, 3)$.
\end{lemma}
\noindent\textbf{Proof of  Lemma~\ref{lemma_W_hat_RE}.} Applying SVD to $\frac{1}{\sqrt{n_t}}(X_t^i)^\top = U_t D_t V_t^\top$, where $U_t \in O^{d\times d}, V_t \in O^{n\times d}, D_t = diag(\sigma_{1,t},...,\sigma_{d,t})$. Let $Q_t := V_t^\top Z_t\Delta_t$, we know $Q_t \sim \mathcal{N}(0, \sigma_z^2\Delta_t^2I_d)$ since $X_t^i, \Delta_t$ are independent to $Z_{t}$, so does $V_t$. Furthermore, we have $\sum_{t=1}^T\frac{1}{\sqrt{n_t}}U_tD_tQ_t \sim \mathcal{N}(0,\sigma_z^2\sum_{t=1}^T\frac{1}{n_t}\Delta_t^2 U_tD_t^2U_t^\top) = \mathcal{N}(0,\sigma_z^2\sum_{t=1}^T\frac{1}{n_t}\Delta_t^2 \hat{\Sigma}_t^i)$ due to task independence. Notice that:
\begin{equation}\label{equ_UD2U}
    (1-2\epsilon) I_d \preceq \hat{\Sigma}_t^i = \frac{1}{n_t}(X_t^i)^\top X_t^i = U_tD_t^2U_t^\top
    \preceq (1+2\epsilon) I_d,\qquad(\text{Assumption}~\ref{assump_Sigma}, (\ref{inequ_Sigma}))
\end{equation}
We immediately have $\sigma_{*}(D_t) \in [\sqrt{1-2\epsilon}, \sqrt{1+2\epsilon}]$. From the density function of multivariate normal distribution, let $\hat{\Gamma}:=\sum_{t=1}^T\frac{1}{n_t}\Delta_t^2\hat{\Sigma}_t^i$ and $\widetilde{\Delta}_t = \frac{\Delta_t}{\sqrt{n_t}}$, then from (\ref{equ_UD2U}), when $\Vert x\Vert_2$ is sufficiently large we have:
\begin{equation}
\begin{split}
    \frac{1}{(2\pi)^{\frac{d}{2}}\Vert\widetilde{\Delta} \Vert_2\sqrt{1+2\epsilon}}\exp(-\frac{1}{2}x^\top x \frac{1}{\Vert\widetilde{\Delta} \Vert_2^2(1+2\epsilon) })
    \geq
    \frac{1}{(2\pi)^{\frac{d}{2}}|\hat{\Gamma}|^{1/2}}\exp(-\frac{1}{2}x^\top\hat{\Gamma}^{-1}x )
\end{split}
\end{equation}
Thus in order to bound the L2 norm of $\sum_{t=1}^T\frac{1}{\sqrt{n_t}}U_tD_tQ_t$ with high probability, we just need to bound the L2 norm of random vectors with distribution $\mathcal{N}(0,\sigma_z^2(1+2\epsilon)\Vert\widetilde{\Delta} \Vert_2^2)$. Let $\xi \sim \mathcal{N}(0,\sigma_z^2(1+2\epsilon)\Vert\widetilde{\Delta} \Vert_2^2)$, like (\ref{inequ_Q}), with probability at least $1 - \frac{\delta}{4}$ we have:
\begin{equation}\label{inequ_xi}
    \Vert \xi\Vert_2^2 \leq \sigma_z^2(1+2\epsilon)\Vert\widetilde{\Delta} \Vert_2^2(2d + 3\ln(\frac{4}{\delta}))
\end{equation}
Then with probability at least $1 - \frac{\delta}{4}$ we have the following inequality for all $\Delta \in \R^{T}$
\begin{equation}
    \begin{split}
        \Vert \hat{W}\Delta\Vert_2 &= \Vert\sum_{t=1}^T((\hat{B}^i)^\top\hat{\Sigma}_{t}^i \hat{B}^i)^{-1}(\hat{B}^i)^\top\hat{\Sigma}_{t}^i B^*w_{t}^*\Delta_t + \sum_{t=1}^T\frac{1}{n_t}((\hat{B}^i)^\top\hat{\Sigma}_{t}^i \hat{B}^i)^{-1}(\hat{B}^i)^\top (X_t^i)^\top Z_t\Delta_t\Vert_2\\
        &\geq |\Vert\sum_{t=1}^T((\hat{B}^i)^\top\hat{\Sigma}_{t}^i \hat{B}^i)^{-1}(\hat{B}^i)^\top\hat{\Sigma}_{t}^i B^*w_{t}^*\Delta_t\Vert_2 - \Vert\sum_{t=1}^T\frac{1}{n_t}((\hat{B}^i)^\top\hat{\Sigma}_{t}^i \hat{B}^i)^{-1}(\hat{B}^i)^\top (X_t^i)^\top Z_t\Delta_t\Vert_2|\\
        &\geq |\frac{1-2\epsilon}{1+2\epsilon}\Vert W^*\Delta\Vert_2 - \frac{1}{1-2\epsilon}\Vert(\hat{B}^i)^\top(\sum_{t=1}^T\frac{1}{n_t} (X_t^i)^\top Z_t\Delta_t)\Vert_2|\\
        &\geq |\frac{1-2\epsilon}{1+2\epsilon}\Vert W^*\Delta\Vert_2 - \frac{1}{1-2\epsilon}\Vert\sum_{t=1}^T\frac{1}{\sqrt{n_t}}U_tD_tQ_t\Vert_2|\\
        &\geq |\frac{0.9}{1.1}\Vert W^*\Delta\Vert_2 - \frac{\sqrt{1.1}}{0.9}\sigma_z\Vert\Delta \Vert_2\sqrt{\frac{(2d + 3\ln(\frac{4}{\delta}))}{\min_t(n_t)}}|, \qquad(\text{Conditions}, (\ref{inequ_xi}))\\
        &\geq |\frac{0.9}{1.1}\sqrt{\kappa}\Vert\Delta\Vert_2 - \frac{\sqrt{1.1}}{0.9\times 4}\sqrt{\kappa}\Vert\Delta \Vert_2|, \qquad(n_t \geq 12 \frac{\sigma_z^2}{\kappa}(d+\ln(\frac{4}{\delta})))\\
        &\geq 0.5\sqrt{\kappa}\Vert\Delta \Vert_2
    \end{split}
\end{equation}
From the definition of RE condition like Assumption~\ref{assump_RE}, we done the proof.
\qed


\begin{lemma}\label{lemma_c_epsilon}
Let $q = \frac{\sqrt{k}R}{\usigma}$ (so $q \geq \|\nu^*\|_1$). If $\gamma \geq \max\{2160sqC_W\Lambda^{-1}, \sqrt{2160sq\kappa\Lambda^{-1}}\}$, then 
\begin{equation}\label{inequ_multi_term}
    \frac{2160}{\gamma}sq  \max\{C_W,\frac{\kappa}{\gamma}\} \leq \|\nu^*\|_1
\end{equation}
\end{lemma}
\noindent\textbf{Proof of  Lemma~\ref{lemma_c_epsilon}.} Just note that if $\gamma \geq \max\{2160sqC_W\|\nu^*\|_1^{-1}, \sqrt{2160sq\kappa\|\nu^*\|_1^{-1}}\}$, then we can prove (\ref{inequ_multi_term}) by direct calculation. Then since $\|\nu^*\|_1 \geq \Lambda$ by definition, we get the result.

\qed

\noindent\textbf{Proof of Theorem~\ref{thm_noise}/\ref{thm_noise_rewrite}.}
Combine Theorem~\ref{thm_nu_hat_bound} and Lemma~\ref{lemma_c_epsilon} and we can figure out the result like (\ref{inequ_practical_L1_UB}).
For the estimation of $\beta_1 = T\beta$, we use the fact that $s \leq k$, $\|\nu\|_2 \leq R/\underline{\sigma}$ and $
\|\nu\|_1 \geq \|\nu\|_2 \geq R/\overline{\sigma}$,
then from the definition of $\gamma$, we can figure out that 
$\beta_1$ should be at least $ \Theta( Tk^3 C_W^6/\underline{\sigma}^6)$.
\qed

\section{Proof of Theorem~\ref{thm_general_cost_optimize}}\label{appendix_task_selection}
First, we rewrite the assumption and theorem formally.
\begin{assumption}\label{assump_f_decrease} (decreasing gradient)
Assume $f_t$ is a piecewise second-order differentiable function, and on each sub-function, it satisfies $f_t \geq 0, \nabla f_t \geq 0, \nabla^2 f_t \leq 0$ and $\nabla f_t(n_{t,1}+n_{t,2}) = \Omega( n_{t,2}^{-2+q})$ for some $q \in (0, 2]$.
\end{assumption}
\begin{remark}
Assumption~\ref{assump_f_decrease} covers a wide range of functions that may be used in practice, including the above example (\ref{fun_saltus_cost}). The last upper bound constraint in Assumption~\ref{assump_f_decrease} shows that we need $\gradient f_t$ to decrease moderately, and it's used for our main theorem in this section. 
\end{remark}

And our main result for Section~\ref{sec_task_selection} is:

\begin{theorem}\label{thm_general_cost_optimize_2}
Let $n_{t,1} \equiv n_1$ for all $t \in [T]$ and assume 
Assumption
~\ref{assump_subG},
\ref{assump_dimension}, \ref{assump_W_fullr},
\ref{assump_bounded_W},
\ref{assump_f_decrease} hold. Without loss of generality, we also assume $R = \Theta(1)$ and $C_W = \Theta(1)$ where $C_W, R$ are defined in Assumption~\ref{assump_bounded_W}.
Then denotes the optimal solution of (\ref{opt_F_simple}) as $(n^*_{[T],2}, \nu^*)$, we have
\begin{equation}\label{equ_propto_n_t}
     n_{t,2}^* = h_t(|\nu^*(t)|)
 \end{equation}
where $h_t$ is a monotone increasing function that satisfies: $c_{t,1}x \leq h_t(x) \leq c_{t,2}x^{2/q}$ where $c_{t,1}, c_{t,2} > 0$ and $q$ defined in Assumption~\ref{assump_f_decrease}. Moreover, we claim A-MTRL algorithm with $n^*_{[T],[2]}$ sampling strategy is at least $k$-sparse task selection algorithm.
\end{theorem}

And we also rewrite the optimization problem (\ref{opt_F_simple}) formally:
\begin{equation}\label{opt_F}
    \begin{aligned}
      &\min_{n_{[T],2}}& g(n_{[T],2})&:= \sum_{t = 1}^T f_t(n_{t,1}+ n_{t,2})&& \\
        &s.t.& c_0(n_{[T],2},\nu)&:=\frac{\varepsilon^2}{C_{ER}\sigma^2k(d+T)} - \sum_{t=1}^T\frac{\nu(t)^2}{n_{t,2} + n_{t,1}}&\geq 0, &\\
        & &c_j(\nu) &:=\sum_{t=1}^Tw_{j,t}^*\nu(t)-(w_{T+1}^*)_j &= 0, &\quad j \in [k]\\
        & &c_m(n_{[T],2}) 
        &:= n_{m,2} &\geq 0, & \quad m \in [T]
    \end{aligned}
\end{equation}
where $C_{ER}>0$ is a constant.


\noindent\textbf{Proof of  Theorem~\ref{thm_general_cost_optimize_2}.} 
Here we note that the main insight for such a theorem is that we want to prove the objective function is concave relative to $\nu$. So we just prove for global second-order differentiable function and it can be easily generalized to the piecewise second-order differentiable function by showing the maintenance of concavity.

\textbf{Step 1: Use KKT conditions to reduce the variable's number}

Firstly we define the Lagrange function:
\begin{equation}
    L(n_{[T],2},\nu) = g(n_{[T],2}) - \lambda_0 c_0(n_{[T],2}) - \sum_{j=1}^k \lambda_j c_j(\nu) - \sum_{m=1}^T\lambda_{m+k} c_m(n_{[T],2})
\end{equation}
Then from KKT conditions we have
\begin{equation}\label{equ_KKT}
\begin{split}
    \left.\frac{\partial L}{\partial n_{t,2}}\right|_{n_{t,2}^*,\nu^*(t)}  &= \nabla f_t(n_{t,1}+n_{t,2}^*) -\lambda_0^* \frac{\nu^*(t)^2}{(n^*_{t,2}+n_{t,1})^2} - \lambda_{t+k,2}^* = 0, \qquad \forall t \in [T]
    \\
    \left.\frac{\partial L}{\partial \nu_t} \right|_{n_{t,2}^*,\nu^*(t)}&= 2\lambda_0^* \frac{\nu^*(t)}{n^*_{t,2}+n_{t,1}} - \sum_{j=1}^k\lambda_j^* w_{j,t}^* = 0 , \qquad \forall t \in [T]\\
    \lambda_0^* &\geq 0, \qquad \lambda_0^*c_0(n_{[T],2}^*,\nu^*) = 0 \\
    \lambda_{m+k}^* &\geq 0, \qquad \lambda_{m+k}^*c_{m}(n_{[T],2}^*) = 0 , \qquad \forall m \in [T]\\
    \end{split}
\end{equation}
Note that when $n_{t,2}^* > 0$, $\lambda_{m+k}^*=0$ and $\nabla f_t(n_{t,1}+n_{t,2}) = \Omega(  n_{t,2}^{-2+q})$.
then from the first equation of (\ref{equ_KKT}) we deduce (\ref{equ_propto_n_t}) and its property immediately. 


Also, with (\ref{equ_propto_n_t}) we can reduce the number of variables of the original problem from $2T$ to $T+1$. To avoid confusion we denote $\alpha = \sqrt{\lambda_0}, \gamma(t):= \nu(t)$ for new optimization problem (\ref{opt_l}). It's clear that if the optimal solution of the original optimization problem (\ref{opt_F}) is $(\nu^*, n_{[T],2}^*)$ and the corresponding lagrange coefficient for the first equality constraint of (\ref{opt_F}) is $\lambda_0^*$, then the optimal solution $(\gamma^*, \alpha^*)$ of the following problem (\ref{opt_l}) is equal to $(\nu^*,\sqrt{\lambda_0^*})$.

\begin{equation}\label{opt_l}
    \begin{aligned}
      &\min_{\gamma, \alpha}& l(\gamma, \alpha)&:= \sum_{t = 1}^T f_t(n_{t,1}+ h_t(\alpha|\gamma(t)|)) \\
        &s.t.& d_0(\gamma, \alpha) &:= \frac{\varepsilon^2}{C_{ER}\sigma^2k(d+T)} - \sum_{t=1}^T\frac{\gamma(t)^2}{h_t(\alpha|\gamma(t)|) + n_{t,1}} = 0 \\
        && d_j(\gamma) &:= \sum_{t=1}^Tw^*_{j,t}\gamma(t) - (w_{T+1}^*)_j = 0, \quad j \in [k]\\
    \end{aligned}
\end{equation}

\textbf{Step 2: The objective function of (\ref{opt_l}) is concave}

From the KKT conditions above we know for any feasible solution $(\gamma, \alpha)$ and any $t \in [S]$, there exist a unique $x_t > 0$ such that $\alpha |\gamma(t)| = \sqrt{\nabla f_t(n_{t,1}+x)} \cdot (n_{t,1} + x)$. Then from the key Lemma~\ref{lemma_opt_concave} we know the objective function of (\ref{opt_l}) is concave relative to $|\gamma(t)|$ for all $t \in [S]$.


\textbf{Step 3: Analyze $\gamma^*$ from the sub-problem of (\ref{opt_l})}

The first equality constraint of the problem (\ref{opt_l}) is non-linear relative to $\gamma$ and $\alpha$, which results that the feasible region of (\ref{opt_l}) having non-linear boundary. This makes it difficult for us to get the closed form of the optimal solution for (\ref{opt_l}). 

Fortunately, the other equality constraints, which are equivalent to $W^*\gamma = w_{T+1}^*$, are not only linear but also have nothing to do with $\alpha$. So we try to find out the optimal solution of sub-problem (\ref{opt_sub}) and connect it to that of (\ref{opt_l}).

\begin{equation}\label{opt_sub}
    \begin{aligned}
      &\min_{\xi}& l(\xi, \alpha)&:= \sum_{t = 1}^T f_t(n_{t,1}+ h_t(\alpha|\xi(t)|)) \\
        &s.t.& 
       D(\xi)&:=W^*\xi - w_{T+1}^* = 0\\
    \end{aligned}
\end{equation}

In (\ref{opt_sub}) $\alpha$ is taken as a given value and $\xi$ plays the same role as $\gamma$ as above. Define $opt(\alpha) : \R \rightarrow \Omega^*$, where $\Omega^*$ is the set of optimal solutions for (\ref{opt_sub}) with given $\alpha$.

Firstly we show that the optimal solution of (\ref{opt_sub}) is $k$-sparse. From step 2 we know $l(\xi)$ is concave for any $|\xi(t)|, t \in [S]$, which means that the region contained by the isosurface of the objective function is concave where the axes are made up of $|\gamma(t)|$ for $t \in [S]$. Consequently, the solutions of the system of linear equations that minimize such a concave function will give out sparse results \cite{Tibshirani1996RegressionSA}.

Secondly, we say the optimal solution of the original optimization problem (\ref{opt_l}) is $k$-sparse. For a non-trivial case, where the algorithm achieves require performance and terminates at the first stage, we know $d_0(\gamma, 0) < 0$, and if $\alpha \rightarrow \infty$, $d_0(\gamma, \alpha) \rightarrow \frac{\varepsilon^2}{C_{ER}\sigma^2k(d+T)} > 0$. Then from continuality of $h_t$ we see that for any $\gamma^*(\alpha) \in opt(\alpha)$, there exist a unique $\alpha_0$ such that $\gamma^*(\alpha_0)$ is a feasible solution for (\ref{opt_l}). On the other hand, every optimal solution $(\gamma^*, \alpha^*) $ of (\ref{opt_l}) should be the optimal solution of sub-problem (\ref{opt_sub}),i.e. it should satisfy $\gamma^* \in opt(\alpha^*)$. Thus $\gamma^*$ is $k$-sparse, and so as $\nu^*$. Therefore A-MTRL with $n^*_{[T],[2]}$ strategy is $k$-sparse task selection algorithm.

\qed


\begin{lemma}\label{lemma_opt_concave}
Assume $f_t, h_t, n_{t,1}$ follow the conditions and results in Theorem~\ref{thm_general_cost_optimize}, $W^* \in \R^{k\times T}$, $w_{T+1}^* \in \R^{k}$. Then if for any feasible solution $(\gamma, \alpha)$ of (\ref{opt_l}), any $t \in [S]$, there exist a unique $x_t > 0$ such that $\alpha |\gamma(t)| = \sqrt{\nabla f_t(n_{t,1}+x)} \cdot (n_{t,1} + x)$, then the objective function of (\ref{opt_l}) relative to $|\gamma(t)|$ is concave for all $t \in [S]$.

\end{lemma}

\noindent{\textbf{Proof of Lemma~\ref{lemma_opt_concave}.}}

Firstly we denote $n_{t,1}$ as $n$ for convenience. Note that from the chain rule:
\begin{equation}\label{equ_l_gamma}
    \frac{\partial l(\gamma, \alpha)}{\partial |\gamma(t)|} = \nabla f_t(n+h_t(\alpha|\gamma(t)|))\cdot \nabla h_t (\alpha|\gamma(t)|)) \cdot \alpha\\
\end{equation}
Clearly $l(\gamma, \alpha)$ is also monotone increasing relative to $|\gamma(t)|$. For the second order of $l(\gamma, \alpha)$ we have:
\begin{equation}
    \frac{\partial^2 l(\gamma, \alpha)}{\partial |\gamma(t)|^2} = \{\nabla^2 f_t(n+h_t(\alpha|\gamma(t)|))\cdot (\nabla h_t (\alpha|\gamma(t)|))^2 +  \nabla f_t(n+h_t(\alpha|\gamma(t)|))\cdot \nabla^2 h_t (\alpha|\gamma(t)|)\}\cdot\alpha^2
\end{equation}
Firstly we need to figure out the relation between the derivative of $h_t$ and $f_t$. From the first equation of (\ref{equ_KKT}) and the definition of $h_t$ we have:
\begin{equation}
    h_t(\sqrt{\nabla f_t(n+x)} \cdot (n + x)) = x
\end{equation}
Since $h_t$ is monotone contineous function, from inverse function theory we have
\begin{equation}
    \nabla h_t(\sqrt{\nabla f_t(n+x)} \cdot (n + x)) = \frac{2\sqrt{\nabla f_t(n+x)}}{(n+x)\nabla^2 f_t(n+x) + 2\nabla f_t(n+x)}
\end{equation}
Let $g(x) := \sqrt{\nabla f_t(n+x)} \cdot (n + x)$, from assumption~\ref{assump_f_decrease} we know $g$ is a continous monotone increasing function and $g \in (0, +\infty)$. Besides, 
from conditions we have that for each $t\in [S]$ there is a unique $x := x_t > 0$ such that $g(x_t) = \alpha |\gamma(t)|$, with which we can simplify the gradient:
\begin{equation}
\begin{split}
    \nabla^2 h_t(\alpha|\gamma(t)|)  &= \nabla^2 h_t(\sqrt{\nabla f_t(n+x)} \cdot (n + x))\\
    &= d(\frac{2\sqrt{\nabla f_t(n+x)}}{(n+x)\nabla^2 f_t(n+x) + 2\nabla f_t(n+x)})/dx \cdot \nabla h_t(\sqrt{\nabla f_t(n+x)} \cdot (n + x))\\
    & = 2\frac{(\nabla^2 f_t(n+x))^2(n+x)-4\nabla^2 f_t(n+x)\nabla f_t(n+x)-2(n+x)\nabla^3 f_t(n+x)\nabla f_t(n+x)}{[(n+x)\nabla^2 f_t(n+x) + 2\nabla f_t(n+x)]^3}
\end{split}
\end{equation}
Denote $h_t^1 := \nabla h_t(\sqrt{\nabla f_t(n+x)} \cdot (n + x)), h_t^2 := \nabla^2 h_t(\sqrt{\nabla f_t(n+x)} \cdot (n + x))$. Thus we have:
\begin{equation}
    \begin{split}
      \frac{1}{\alpha^2}\frac{\partial^2 l(\gamma, \alpha)}{\partial |\gamma(t)|^2} &= \nabla^2 f_t(n+x) 
      (\nabla h_t(\sqrt{\nabla f_t(n+x)} (n + x)))^2 +  \nabla f_t(n+x) \nabla^2 h_t(\sqrt{\nabla f_t(n+x)} (n + x))\\
      &= h_t^1 \cdot \frac{\sqrt{\nabla f_t(n+x)}(n+x)}{[(n+x)\nabla^2 f_t(n+x) + 2\nabla f_t(n+x)]^2}\cdot \{ 3(\nabla^2 f_t(n+x))^2 - 2 \nabla^3 f_t(n+x)\nabla f_t(n+x)\}\\
      &= 2\nabla f_t(n+x)(n+x) \cdot \frac{3(\nabla^2 f_t(n+x))^2 - 2 \nabla^3 f_t(n+x)\nabla f_t(n+x)}{[(n+x)\nabla^2 f_t(n+x) + 2\nabla f_t(n+x)]^3}\\
      &= 2\nabla f_t(n+x)(n+x) \cdot q(x), \qquad (q(x):=\frac{3(\nabla^2 f_t(n+x))^2 - 2 \nabla^3 f_t(n+x)\nabla f_t(n+x)}{[(n+x)\nabla^2 f_t(n+x) + 2\nabla f_t(n+x)]^3})
    \end{split}
\end{equation}

So if $q(x) < 0$ holds for all $x > 0$, we finish the proof. First we assume that $\nabla f_t(y) = \frac{A_t}{(B_t + y)^\delta}$ where $A_t > 0, B_t \geq 0$ and $\delta \in [0, 2-q)$. Then
\begin{equation}\label{equ_qx}
\begin{split}
    q(x) = \frac{3\frac{\delta^2 A_t^2}{(n+x+B_t)^{2\delta+2}} - 2\frac{\delta(\delta+1) A_t^2}{(n+x+B_t)^{\delta+3 + \delta}} }{\frac{2A_t}{(n+x+B_t)^{\delta}} - \frac{\delta A_t (n+x)}{(n+x+B_t)^{\delta+1}}}
    = \frac{A_t}{(n+x+B_t)^{\delta+1}}\cdot \frac{\delta(\delta - 2)}{2B_t + (2-\delta)(n+x)}
    \end{split}
\end{equation}

Since $n+x > 0$ and $0 \leq \delta < 2$, we have $q(x) < 0, \forall x > 0$. 
%
Besides, due to the fact that $\nabla f_t$ is monotone decreasing and non-negative, together with Assumption~\ref{assump_f_decrease} and $n > 0$, we can find $\delta_i \in [0, 2-q), A_{t, i} > 0, B_{t,i} \geq 0 $ for $i = 1, 2$ such that $\frac{A_{t,1}}{(B_{t,1} + x + n)^{\delta_1}} \leq \nabla f_t(x + n) \leq \frac{A_{t,2}}{(B_{t,2} + x + n)^{\delta_2}}$. So $q(x) < 0$ holds for any $\nabla f_t$ that satisfies Assumption~\ref{assump_f_decrease}.

\qed

\begin{remark}
If $\delta$ in (\ref{equ_qx}) is in $(0, 2)$, then the optimization problem (\ref{opt_F}) is not computable. 
\end{remark}

\section{Supplements to the Experiments Section}

\subsection{Explanation of k-task selection scenario}\label{sec_app_k_task_illu}
We provide an illustration of our intuition for the k-task selection scenario in Section~\ref{sec_experiment}. We emphasize that the specific choice of the cost function is not critical in such a scenario, since solving the exact optimization problem (Eqn.~\ref{opt_F_simple}) can be computationally challenging. For instance, the cost functions could correspond to $L_p$-minimization ($0 \leq p < 1$) solutions of the relation equation $W^*\nu = w^*_{T+1}$, which is known to be NP-hard.

To address this challenge, as discussed in Theorem~\ref{thm_general_cost_optimize}, we employ L1-A-MTRL as an approximation to the optimal solution of (\ref{opt_F_simple}). This approach is justified by the fact that 
the time complexity for solving the approximate solution of (\ref{opt_F_simple}) using L1-A-MTRL with relative accuracy $\delta$ is just $\text{poly}(T)\ln(T/\delta)$ from \cite{cohen2021solving}, and
the $L_1$-minimization solution is also k-sparse. 
Therefore, in cost-sensitive scenarios, our main focus is on addressing the question: "How well can active multi-task representation learning algorithms perform when no more than k tasks are available for further sampling?" This leads us to the setting of the k-task selection scenario.

\subsection{Details of Algorithm Implementation.}
\label{appendix_exp_algo}

In practice, $\hat{W}$ and $\hat{w}_{T+1}$ may differ at different epochs after the model converges due to the noise of data points. So to enhance the stability of $\hat{\nu}$, we calculate $\hat{\nu}$ at every epoch in the last $20$ rounds and take their average as the final reference to calculate $n_{[T]}$ for both our algorithm and baselines, while the total number of epochs at each stage is no less than 2000. 
For full tasks scenario,
note that L2-A-MTRL\cite{chen2022active} utilize the iterative L2-A-MTRL algorithm with 4 stages to optimize the model
we also run our algorithm iteratively with 4 stages for comparison, and the detailed procedure for multi-stage learning is in \cref{alg:L1_A_MTRL_multi}. 
We mention that Chen's method requires multiple stages but we allow both single-stage (\cref{alg:L1_A_MTRL}) and multi-stage (\cref{alg:L1_A_MTRL_multi}) versions.

Here we set $\underline{N}=100$. We sample $500$ data from the target task, while at the final stage, we sample around 30000 to 40000 data from the source tasks. 
For k-task selection scenario, we run the algorithm with 2 stages. Here we set $\underline{N}=40$. We sample $200$ data from the target task and around 12000 data from the source tasks.

\begin{algorithm}[tb]
   \caption{Multi-Stage L1-A-MTRL Method}
   \label{alg:L1_A_MTRL_multi}
\begin{algorithmic}
   \STATE {\bfseries Input:} confidence $\delta$, representation function class $\Phi$, stage number $S$, scaling $L>1$, 
   minimum singular value $\underline{\sigma}$
   \STATE Initialize $\underline{N} = \beta_1/T$ from (\ref{equ_beta}) and $\hat{\nu}^1 = [1, 1, ...]$.
   \FOR{$i=1$ {\bfseries to} S}
   \STATE Set $n_{t}^i = \max\{\beta_i |\hat{\nu}^i(t)|\cdot \|\hat{\nu}^i\|_1^{-1}, \underline{N}\}$.
   \STATE For each task $t$, draw $n_t$ i.i.d samples from the offline dataset denoted as $\{X_t^i, Y_t^i\}_{t=1}^T$
   \STATE Estimate $\hat{\phi}^i, \hat{W}^i$ on the source tasks with Eqn.(\ref{equ_train_on_source}) 
   \STATE Estimate $\hat{w}^i_{T+1}$ on the target task with Eqn.(\ref{equ_train_on_target})
   \STATE Estimate $\nu^{i+1}$ by
   Lasso Program (\ref{equ_Lasso})
   \STATE Set $\beta_{i+1} = \beta_i \cdot L$
   \ENDFOR
\end{algorithmic}
\end{algorithm}

\subsection{How to choose $\lambda_k$}\label{appendix_choose_lambdak}
Determining the optimal value of $\lambda_k$ requires additional knowledge of $\usigma = \sigma_{\min}(W^*)$, which are dataset-dependent prior parameters. To address this, we explore two approaches to determine $\lambda_k$ in our experiments:

\begin{itemize}
\item Tuning way: We roughly tune $\lambda_k$ exponentially for the 2-phase L1-A-MTRL approach (\cref{alg:L1_A_MTRL}). And to further obtain the optimal $\lambda_k$, we can utilize grid search to find better $\lambda_k$. Once we identify a good $\lambda_k$, we can run the multi-phase L1-A-MTRL algorithm (\cref{alg:L1_A_MTRL_multi}) using that $\lambda_k$ and a larger $N_{tot}$ to achieve improved results.
\item Lazy way: Alternatively, we can simply choose a very small value for $\lambda_k$, such as $10^{-10}$, for our algorithm.
\end{itemize}

To provide a clearer illustration of the first approach, we apply the 2-phase L1-A-MTRL on the \textit{identity\_9} dataset in full task scenarios, where $k=50$ and $T=150$. In the first phase, each source task is assigned $\underline{N}=100$ data points, and in the second phase, the total budget for the source data is $N_{tot}=33k$. The results are presented in Table~\ref{table:lambda_error}.

\begin{table}[htbp]
\centering
\caption{The relevance between $\lambda_k$ and the second-stage test error on the target task \textit{identity\_9}.}
\begin{tabular}{@{}lccccccccccc@{}}
\toprule
$\lambda_k$ & $1.0$ & $10^{-1}$ & $10^{-2}$ & $10^{-3}$ & $10^{-4}$ & $2\times 10^{-4}$ & $10^{-5}$ & $10^{-6}$ & $10^{-8}$ & $10^{-10}$ & $10^{-16}$ \\
\midrule
\textbf{Error} & 0.0691 & 0.0690 & 0.0703 & 0.0694 &\textbf{0.0561} & 0.0570 & 0.0655 & 0.0655 & 0.0633 & 0.0631 & 0.0625 \\
\bottomrule
\end{tabular}
\label{table:lambda_error}
\end{table}

The optimal value for $\lambda_k$ is approximately $10^{-4}$. Additionally, we observe that, except for the terms $10^{-4}$ and $2\times 10^{-4}$, the target error decreases as $\lambda_k$ decreases. For other target tasks, although we don't find an optimal $\lambda_k$ similar to that of \textit{identity\_9}, we consistently observe that smaller values of $\lambda_k$ lead to better performance for L1-A-MTRL.
We think this phenomenon can be attributed to our Theorem~\ref{thm_worst_target_ER}, which considers a worst-case scenario where the noise may be significant. However, in practice, smaller values of $\lambda_k$ are often sufficient to control the noise. Furthermore, since smaller values of $\lambda_k$ result in a smaller bias when solving the Lasso program, L1-A-MTRL with small $\lambda_k$ consistently exhibits good performance.

Therefore, to save time and resources, we adopt the lazy way instead of the tuning way in the experiments presented in this paper. We set $\lambda_k = 10^{-10}$, and the empirical results demonstrate that L1-A-MTRL with such a small value of $\lambda_k$ still achieves excellent performance.

\subsection{Additional Experiments on Sampling Budgets}
\label{sec_add_exp_N}
To better show the empirical difference of the sampling budget in the experiments of MNIST-C, we consider the full task scenario (mentioned in Sec.~\ref{sec_experiment}) 
and evaluate the model performance by utilizing the 5-epoch L1-A-MTRL (Algorithm~\ref{alg:L1_A_MTRL_multi}) with fixed minimum sampling data from every source task $\underline{N}=100$ and increasing total sampling number $N_{tot}$. 
Due to the limited time and resources, we randomly select two target tasks \textit{shear\_1} and \textit{identity\_9}, and obtained the results in Table~\ref{table:shear_1}.

From Table~\ref{table:shear_1} we find that to achieve accuracy higher than 95\% on the \textit{shear\_1} target task, P-MTRL (passive sampling) requires more than $86k$ source data, L2-A-MTRL\cite{chen2022active} requires about $61k$ source data and L1-A-MTRL just requires about $33k$ source data. Since at the later phase, we can reuse the evenly sampled data ($T\underline{N}=15k$ in total) from the first phase, L1-A-MTRL just requires labeling additional $18k$ source data at the later phase to achieve 95\% accuracy, while L2-A-MTRL requires approximately $46k$ extra data, and P-MTRL requires no less than $71k$ extra data.
Similar results apply to \textit{identity\_9}
. To achieve an accuracy above $93.7\%$ on the \textit{identity\_9} target task, P-MTRL requires more than $95k$ source data, L2-A-MTRL\cite{chen2022active} requires about $61k$ source data, while L1-A-MTRL requires only about $33k$ source data. The above results illustrate the effectiveness of our L1-A-MTRL algorithm.

\begin{table}[h]
\centering
\caption{Test error on the target task \textit{shear\_1} and \textit{identity\_9} 
 with different $N_{tot}$.}
\begin{tabular}{@{}lccccc@{}}
\toprule
\textit{shear\_1} & \multicolumn{5}{c}{$N_{\text{tot}}$} \\
\cmidrule(l){2-6}
\textbf{Algorithms} & \textbf{15000} & \textbf{32850} & \textbf{44000} & \textbf{60700} & \textbf{86000} \\
\midrule
P-MTRL & 0.0544 & 0.0538 & 0.0536 & 0.0520 & 0.0518 \\
L2-A-MTRL(\citet{chen2022active}) & 0.0544 & 0.0511 & 0.0519 & 0.0494 & 0.0488 \\
L1-A-MTRL(\textbf{Ours}) & 0.0544 & \textbf{0.0496} & \textbf{0.0478} & \textbf{0.0442} & \textbf{0.0428} \\
\bottomrule
\end{tabular}
\label{table:shear_1}
\vspace{1pt}
\centering
\begin{tabular}{@{}lccccc@{}}
\toprule
\textit{identity\_9} & \multicolumn{5}{c}{$N_{\text{tot}}$} \\
\cmidrule(l){2-6}
\textbf{Algorithms} & \textbf{15000} & \textbf{33000} & \textbf{43800} & \textbf{60900} & \textbf{95400} \\
\midrule
P-MTRL & 0.0932 & 0.0834 & 0.0778 & 0.0738 & 0.0652 \\
L2-A-MTRL(\citet{chen2022active}) & 0.0932 & 0.0909 & 0.0638 & 0.0627 & 0.0621 \\
L1-A-MTRL(\textbf{Ours}) & 0.0932 & \textbf{0.0631} & \textbf{0.0620} & \textbf{0.0605} & \textbf{0.0551} \\
\bottomrule
\end{tabular}
\end{table}

\end{document}